\pdfoutput=1
\documentclass[11pt]{article}

\usepackage{acl}

\usepackage{times}
\usepackage{latexsym}

\usepackage[T1]{fontenc}
\usepackage[utf8]{inputenc}

\usepackage{microtype}

\usepackage{graphicx}
\usepackage{multirow}
\usepackage{tabularx}
\usepackage{tabulary}
\usepackage{siunitx}
\usepackage{fullpage}
\usepackage{makecell}
\usepackage{booktabs}
\usepackage{arydshln}
\usepackage{float}
\usepackage{nicematrix}
\usepackage{xspace,mfirstuc,tabulary}
\usepackage{cleveref}
\usepackage{bm}
\usepackage{algpseudocode}
\usepackage{algorithm}
\usepackage{siunitx}
\usepackage{enumitem}

\DeclareMathOperator*{\argmax}{arg\,max}

\newcolumntype{C}{>{\centering\arraybackslash}X}
\newcommand{\modelcolwidth}{4.55 cm}
\newcommand{\ourspace}{0.085cm}  % space before our results

\newcommand{\topleftdesc}[2]{\multirow{#2}{*}{ \rotatebox{90}{ \textbf{#1} }  } & }
\newcommand{\sidedescwidth}{0.2cm}

\title{Label Semantics for Few Shot Named Entity Recognition}

\author{Jie Ma$^1$ \ \ \bf{Miguel Ballesteros}$^1$ \ \ \bf{Srikanth Doss}$^1$ \ \ \bf{Rishita Anubhai}$^1$ \\ \bf{Sunil Mallya}$^1$\thanks{ ~ Work done while at AWS AI Labs.} \ \ \bf{Yaser Al-Onaizan}$^1$\footnotemark[1] \ \ \bf{Dan Roth}$^{1,2}$ \\ $^1$AWS AI Labs \\ $^2$Computer and Information Science, University of Pennsylvania \\ \texttt{\{jieman, ballemig, srikad, ranubhai, drot\}@amazon.com} \\ \texttt{mallya16@gmail.com, onaizan2000@yahoo.com}}

\begin{document}
\maketitle
\begin{abstract}
We study the problem of few shot learning for named entity recognition. Specifically, we leverage the semantic information in the names of the labels as a way of giving the model additional signal and enriched priors. We propose a neural architecture that consists of two BERT encoders, one to encode the document and its tokens and another one to encode each of the labels in natural language format. Our model learns to match the representations of named entities computed by the first encoder with label representations computed by the second encoder. The label semantics signal is shown to support improved state-of-the-art results in multiple few shot NER benchmarks and on-par performance in standard benchmarks. Our model is especially effective in low resource settings.
% This additional label semantic signal makes the proposed model capable of achieving the best results reported in few shot benchmarks by requiring even less data.
% \dr{The last sentence is too weak. How about: Within our proposed architecture, the label semantics signal is shown to support improved state-of-the-art results in multiple few short NER benchmarks, and to be especially effective in low resource settings.}
\end{abstract}

\section{Introduction}
% \dr{Check here for some guidelines on how to structure the introduction: \url{https://www.seas.upenn.edu/~cis620dr/resources.html} (note: these are only guidelines, but provide a good starting point)}
Named entity recognition (NER) seeks to locate named entity spans in unstructured text and classify them into pre-defined categories such as PERSON, LOCATION and ORGANIZATION \cite{tjong-kim-sang-de-meulder-2003-introduction}. As a fundamental natural language understanding task, NER often serves as an upstream component for more complex tasks such as question answering \cite{molla-etal-2006-named}, relation extraction \cite{chan-roth-2011-exploiting} and coreference resolution \cite{clark-manning-2015-entity}. However, building an accurate NER system has traditionally required large amounts of high quality annotated in-domain data \cite{lison-etal-2020-named, DBLP:journals/corr/abs-2010-01677}. This usually involves well defined annotation guidelines and training of annotators, which requires rich domain knowledge and can be prohibitively expensive \cite{DBLP:journals/corr/abs-2012-14978}.
%\mb{is there a paper we can cite regarding this claim?}\jm{couple of citations added for the claim above}

Few shot learning (FSL) \cite{vinyals2017matching, finn2017modelagnostic, snell2017prototypical} aims at performing a task using only very few annotated examples (i.e. support set). % For NER, leveraging few shot techniques can help build entity recognizers with only a handful of annotated examples for each entity type. Previous studies mainly focus on approaches to transfer knowledge from high resource domains to a new low resource domain \cite{DBLP:journals/corr/abs-2106-15167, athiwaratkun2020augmented, hou-etal-2020-shot}.

Similarity-based methods, such as prototypical networks, are extensively studied and show great success for FSL \cite{vinyals2017matching, snell2017prototypical, yu-etal-2018-diverse, hou-etal-2020-shot}. The core idea is to classify input examples from a new domain based on their similarities with representations of each class in the support set. 
% Usually the similarity functions are learned from high resource domains.
%{Some approaches for NER commonly use conditional random field (CRF) \cite{McDonald2005IdentifyingGA} to capture transitions between predicted tags in sequence tagging \cite{yang2020simple, hou-etal-2020-shot}. }
% \ra{Previous studies in FSL focus on similarity based  methods on class representations. \cite{DBLP:journals/corr/abs-2106-15167, athiwaratkun2020augmented, hou-etal-2020-shot}. \cite{vinyals2017matching, snell2017prototypical, yu-etal-2018-diverse}. The core idea here is to classify new input examples based on their similarities with representations of each class in the support set. %{Usually the similarity functions are learned from high resource domains.}
% Previous work on traditional NER approaches on fully supervised large datasets commonly use conditional random fields (CRF) \cite{McDonald2005IdentifyingGA} to capture transitions between predicted tags in sequence tagging.}
These methods do not utilize the semantics of label names and usually represent labels by directly averaging the embedding of support set examples, oversimplifying the learning of label representations. The main premise of our work is that label names carry meaning that our models can induce from data; the labels are themselves words that appear in text in various contexts and are thus semantically related to other words that appear in text, and this relatedness can be leveraged.
%are usually semantically related to each word in the text, so we leverage the similarities between words and labels. 
For example, the representation of ``Lionel Messi'' is more similar %related 
to that of PERSON than to the representations of %it should be to 
LOCATION or DATE when similar priors are used for labels and words or phrases.

In this work, we propose a neural architecture that uses two separate BERT-based encoders \cite{devlin2019bert} to leverage semantics of label names for NER.\footnote{Our model is similar to the two-tower model widely adopted in question answering \cite{DBLP:journals/corr/abs-2004-04906}, recommender systems \cite{DBLP:journals/corr/abs-2102-06156} and entity linking \cite{logeswaran-etal-2019-zero,DBLP:journals/corr/abs-2010-11333}.} One encoder (a) is used to encode the document and its words while the other encoder (b) is used to encode label names (e.g. PERSON, LOCATION etc.). The model is trained to match word representations from encoder (a) with label representations from encoder (b), and assign a label for each word by maximizing the similarity. We also experiment by replacing the BERT label encoder with GloVe embeddings \cite{pennington2014glove} as a simplified architecture.% We pre-finetune the model on source domain and then continue finetuning it on target domain.

% We perform extensive experiments in multiple NER datasets from different domains. We find that our model significantly outperforms strong few-shot baselines in low resource setting and achieve the same level of performance as state-of-the-art models in high resource settings. Furthermore, we also show that that our model is robust to different semantic variation of labels. % (spelling, semantics, etc)

We report experimental results in multiple NER datasets from different domains. We summarize our contribution as follows: 
\begin{itemize}
    \item We propose a simple and effective model architecture that leverages label semantics for NER.
    \item We show that the proposed model is particularly effective in low resource settings and gives on-par results with the state-of-the-art models in high resource settings. 
    \item We achieve a new state-of-the-art in multiple few shot NER benchmarks. Specifically, our model outperforms prior work by 1.2 to 6.6 F1 points on CoNLL'03, WNUT'17, JNLPBA, NCBI-disease and I2B2'14 datasets on various few shot shots settings (\S \ref{exp:results}).
    \item We show that the proposed model is robust to variations of label names and that it is able to differentiate semantically similar labels.
\end{itemize}

\begin{figure*}
  \includegraphics[width=\textwidth]{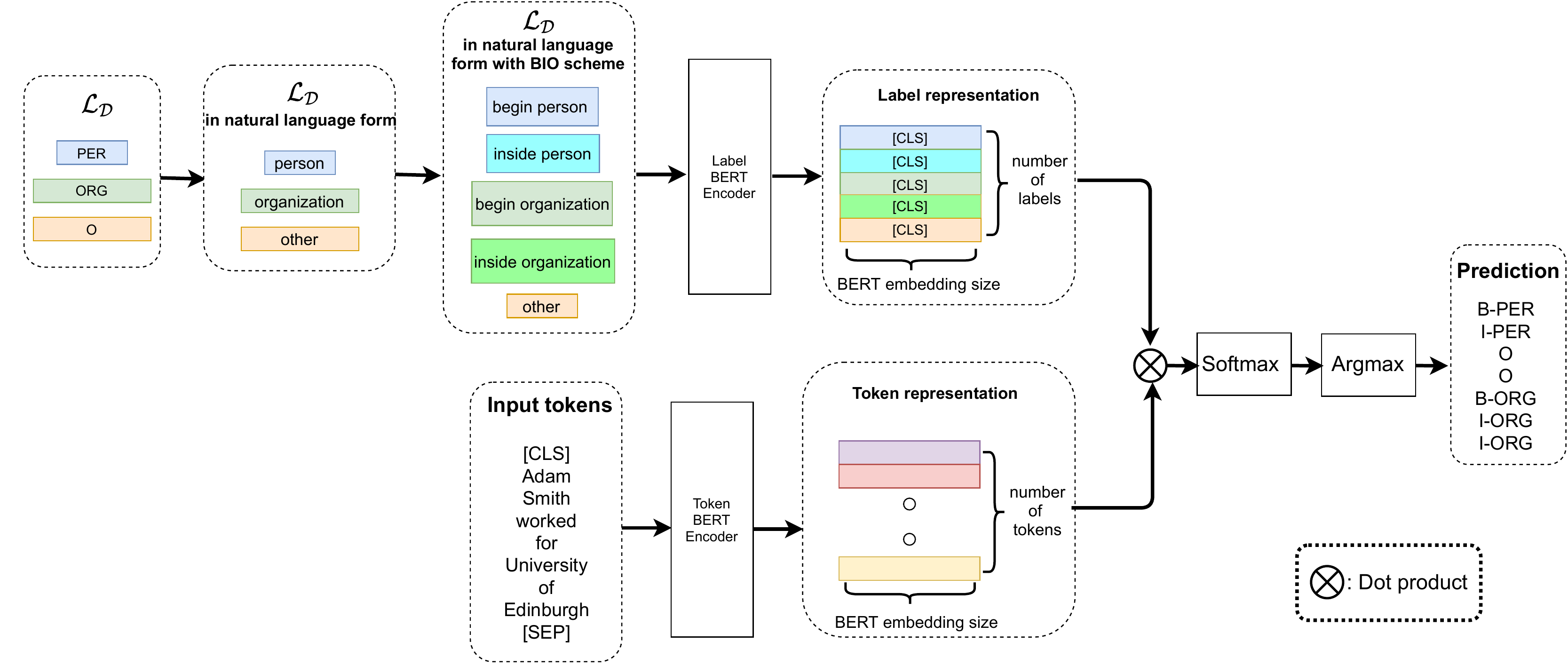}
  \caption{The architecture of our NER model. The diagram shows how representation of labels and tokens are produced, and how we use them to calculate final model prediction. The top part of the figure shows how labels are encoded; the bottom part of the figure shows how sentence are encoded.}
  \label{fig:model_arch}
\end{figure*}

\section{Model}
We present our NER model. As shown in Figure \ref{fig:model_arch}, it consists of two BERT-based encoders where one encoder is used to encode the document and its tokens and the other to encode labels. We formalize the differences between datasets used in our experimentation (\S \ref{model:datasets}), then present how two BERT-based encoders (and the modification with GloVe-based encoder for labels) are used to leverage semantics in labels for NER (\S \ref{model:architecture}). Finally we discuss the training procedure (\S \ref{model:training}) and how labels are represented (\S \ref{model:label_rep}).

%\subsection{Notation} \label{model:notation}
%\mb{Jie: do we use this notation about the task anywhaere in the paper? If not, why do we have it?}
%We define a sentence as sequence of tokens $\boldsymbol{t} = (t_{1}, t_{2}, ..., t_{n})$ and corresponding NER labels for each token as $\boldsymbol{y} = (y_{1}, y_{2}, ..., y_{n})$. A dataset is defined as a set of $(\boldsymbol{t}, \boldsymbol{y})$ pairs: $\mathcal{D} = \{(\boldsymbol{t}^{(i)}, \boldsymbol{y}^{(i)})\}_{i=1}^{N_{D}}$, where $N_{D}$ is number of sentences in the dataset. Each dataset is associated with a specific set of labels $\mathcal{L_{D}} = \{\ell_{1}, \ell_{2}, ..., \ell_{N_{L}}\}$, where $N_{L}$ is number of unique labels. Given a training dataset $\mathcal{D}$ and a query sentence $\boldsymbol{t}$, the task of NER is to find the optimal sequence of labels $\boldsymbol{y}^{*}$ for $\boldsymbol{t}$:

%\[\boldsymbol{y}^{*} = \argmax_{\boldsymbol{y}}\ p(\boldsymbol{y} | \boldsymbol{t}, \mathcal{D}, \mathcal{L_{D}})\]

\subsection{Source and Target Datasets}
\label{model:datasets}
For few shot NER, we use a setup similar to meta-learning. We first train our models on \textbf{source datasets} $\{\mathcal{D}_{1}^{S}, \mathcal{D}_{2}^{S},...\}$, then evaluate the model on unseen few shot \textbf{target datasets} $\{\mathcal{D}_{1}^{T}, \mathcal{D}_{2}^{T}, ...\}$ with or without finetuning. Each target dataset only contains a few examples and a different taxonomy of labels compared to the source datasets.%\mb{we should refer to this paragraph in the Experimental Setup section and explain what is SOURCE and what iS TARGET dataset}
%\ra{Right now this is painfully similar to metalearning setup. Can we use the same notations as them and in case we can't let's atleast add a reference for it.}

\subsection{Architecture} \label{model:architecture}
We use two BERT-based encoders as shown in Figure~\ref{fig:model_arch}: a BERT document encoder and a BERT label encoder (we also experiment with GloVe embeddings as label encoder, described in \S \ref{experiment:glove-as-label-encoder}).  Like the traditional NER models \cite[inter alia]{carreras-etal-2003-learning,DBLP:journals/corr/abs-1103-0398, lample-etal-2016-neural}, we predict the label of each token with BIO scheme.\footnote{Each token is predicted as B-entity\_type, I-entity\_type or O, indicating the token is at the beginning, inside or outside of the entity\_type.} For each token we get an embedding $e$ from the first BERT document encoder. For the unique set of labels $\mathcal{L_{D}}$ associated with dataset $\mathcal{D}$, we apply three steps to get the representations: First, we manually convert the label names to their natural language forms, e.g. ``PER'' to ``person'', ``ORG'' to ``organization'' etc. Second, we convert each of the label names to BIO scheme, in the form of natural language, e.g. ``person'' to ``begin person'' or ``inside person''. Finally, we use the second BERT label encoder to embed each of the labels in natural language BIO scheme. We compute the BERT [CLS] token embedding as the representation for the corresponding label. We form a label vector $\boldsymbol{b}$ of all label embeddings $b_i$ for all $i$ in $\{1, 2, ..., {2 \times N_{L} - 1}\}$ \footnote{Each of the $N_L$ labels are converted to BIO scheme except ``O''/``other'', thus it is $2 \times N_{L}-1$ embeddings in total.}. The label encoder acts like a lookup table for label embeddings. Finally, to find the most appropriate label for this token, we use:

\[y = \argmax_i\ \text{softmax}(e \cdot \boldsymbol{b})\]

%Given a sentence $\boldsymbol{t}$ from dataset $\mathcal{D}$, we first use the BERT document encoder to embed all tokens, where embedding of token $t_i$ is denoted as $b^{t}_i$. For the unique set of labels $\mathcal{L_{D}}$ associated with dataset $\mathcal{D}$, we apply three steps to get the representations: First, we manually convert the label names to their natural language forms, e.g. ``PER'' to ``person'', ``ORG'' to ``organization'' etc. Second, we convert each of the label names to BIO scheme, in the form of natural language, e.g. ``person'' to ``begin person'' or ``inside person''. Finally, we use the second BERT label encoder to embed each of the labels in natural language BIO scheme from second step and use the encoder to compute the [CLS] token representation for the corresponding label. We denote these label embeddings as ${b}^{l} = \{b^{l}_{1}, b^{l}_{2}, ..., b^{l}_{2 \times N_{L} - 1}\}$.\footnote{Each of the $N_L$ labels are converted to BIO scheme except ``O''/``other'', thus it is $2 \times N_{L}-1$ embeddings in total.} In this way the BERT label encoder acts like a lookup table for label embeddings.

%With the embedding of a given token $t_{i}$ as $b_{i}^{t}$, and the embedding of all labels as ${b}^{l}$, we calculate dot product between $b_{i}^{t}$ and ${b}^{l}$, followed by a softmax to produce the final prediction for each token:

%\[y_{i} = \argmax\ \text{softmax}(b_{i}^{t} \cdot {b}^{l})\]

\subsection{Training} \label{model:training}
Comparing with prior work on neural architectures for NER, our model does not require a new randomly initialized top layer classifier for a new dataset with new unseen label names. Instead, we generate label representations from the BERT label encoder. We hypothesize that this is beneficial because it prevents the model from forgetting priors since no parameters are dropped or randomly initialized for different datasets. 

We propose a simple two stage training procedure. In the first stage, we pre-finetune our model on the mix of all source datasets (which usually have different label set taxonomies), then we finetune the trained model on the target dataset. This process is also known as pre-finetuning \cite{DBLP:journals/corr/abs-2101-11038} and finetuning. For scenarios where no source datasets are available, we simply skip the first stage. During model training time, both encoders are updated for every iteration at both stages, which helps to align the token embedding space and the label embedding space.

During inference time, the learned label encoder is only required to produce label representations once. This is because the label representations may be cached and the label encoder is no longer needed to recompute representations. Our model is therefore not introducing additional memory overhead (since label encoder is removed) or latency overhead (since label representation is cached).

\subsection{Label Representation} \label{model:label_rep}
Given that our label encoder is based on BERT and contains the priors from pretraining, our architecture allows any textual form as input for the generation of label representations. In order to make our results comparable with previous studies, we use only the natural language form of label names for our primary results. We discuss more label representations in Appendix \ref{appendix:contextual_label_rep}.

\begin{table*}[!ht]
\centering
\small
\setlength{\tabcolsep}{5pt}
\vspace{-.3cm}

\begin{tabularx}{\textwidth}{p{\sidedescwidth} p{\modelcolwidth} *{5}{C}}
    \topleftdesc{CoNLL-2003}{11}
     & {\bf 1 Shot} & {\bf 5 Shot} & {{\bf 20 Shot}} & {{\bf 50 Shot}} & {{\bf Full Dataset}} \\
\toprule
 & TransferBERT & 44.8 $\pm$ 15.0 & 66.9 $\pm$ 6.7 & 77.5 $\pm$ 1.2 & 82.0 $\pm$ 1.1 & 91.3 $\pm$ 0.2 \\
 & Prototypical Network & 7.5 $\pm$ 2.6 & 11.5 $\pm$ 5.6 & 18.6 $\pm$ 7.5 & 16.3 $\pm$ 2.7 & N/A \\
 & WPN-CRF & 56.26 $\pm$ 9.1 & 67.7 $\pm$ 4.4 & 67.4 $\pm$ 2.0 & 69.0 $\pm$ 1.7 & N/A \\
% &  L-TapNet+CDT & x & x &  x & x & x \\
 &  Struct NN shot & 63.7 $\pm$ 3.7 & 70.0 $\pm$ 3.0 & 73.1 $\pm$ 1.9 & 75.7 $\pm$ 1.8 & N/A \\
 &  TANL & 54.7 $\pm$ 9.4 & 65.6 $\pm$ 3.8 & 71.0 $\pm$ 2.4 & 74.4 $\pm$ 1.9 & \bf 91.7 $\pm$ 0.4 \\ 
    [\ourspace]
    \cmidrule(lr){2-7}
 &  Our model - GloVe & 63.1 $\pm$ 6.9 & 73.5 $\pm$ 2.4 & 78.3 $\pm$ 1.1 & 82.0 $\pm$ 1.5 & 91.6 $\pm$ 0.2 \\
 &  Our model - BERT & \bf 68.4 $\pm$ 6.7 & \bf 76.6 $\pm$ 2.1 & \bf 79.7 $\pm$ 1.1 & \bf 83.1 $\pm$ 1.2 & 91.5 $\pm$ 0.2 \\ 
\end{tabularx}

\begin{tabularx}{\textwidth}{p{\sidedescwidth} p{\modelcolwidth} *{5}{C}}
    \topleftdesc{WNUT-2017}{10} \\
     % & {\bf 1 Shot} & {\bf 5 Shot} & {{\bf 20 Shot}} & {{\bf 50 Shot}} & {{\bf Full Dataset}} \\

\toprule
 & TransferBERT & 27.6 $\pm$ 6.8 & 35.2 $\pm$ 3.4 & 40.9 $\pm$ 1.6 & 42.5 $\pm$ 1.2 & 44.0 $\pm$ 0.2 \\
 & Prototypical Network & 1.7 $\pm$ 1.2 & 2.1 $\pm$ 1.0 & 2.7 $\pm$ 1.6 & 3.5 $\pm$ 1.7 & N/A \\
 & WPN-CRF & 23.1 $\pm$ 2.8 & 29.9 $\pm$ 3.2 & 32.9 $\pm$ 1.2 & 33.2 $\pm$ 1.1 & N/A \\
 % &  L-TapNet+CDT & x & x &  x & x & x \\
 &  Struct NN shot & 31.1 $\pm$ 6.4 & 33.2 $\pm$ 2.0 & 30.8 $\pm$ 2.2 & 31.8 $\pm$ 1.8 & N/A \\
 &  TANL & 25.6 $\pm$ 6.3 & 33.3 $\pm$ 4.4 & 34.1 $\pm$ 2.1 & 34.4 $\pm$ 2.4 & 45.2 $\pm$ 0.6 \\ 
    [\ourspace]
 \cmidrule(lr){2-7}
 &  Our model - GloVe & 36.6 $\pm$ 2.4 & 39.6 $\pm$ 1.9 & 42.5 $\pm$ 1.3 & 43.0 $\pm$ 1.1 & \bf 45.7 $\pm$ 0.6 \\
 &  Our model - BERT & \bf 38.3 $\pm$ 1.7 & \bf 40.8 $\pm$ 2.1 & \bf 42.7 $\pm$ 1.1 & \bf 43.3 $\pm$ 0.8 & 45.0 $\pm$ 0.6 \\ 
\end{tabularx}

\begin{tabularx}{\textwidth}{p{\sidedescwidth} p{\modelcolwidth} *{5}{C}}
    \topleftdesc{JNLPBA}{10} \\
     % & {\bf 1 Shot} & {\bf 5 Shot} & {{\bf 20 Shot}} & {{\bf 50 Shot}} & {{\bf Full Dataset}} \\

\toprule
 & TransferBERT & 26.6 $\pm$ 7.8 & 40.3 $\pm$ 2.8 & 53.2 $\pm$ 2.9 & 59.7 $\pm$ 1.3 & 71.0 $\pm$ 0.5 \\
 & Prototypical Network & 2.1 $\pm$ 1.5 & 4.0 $\pm$ 3.2 & 6.8 $\pm$ 3.6 & 5.7 $\pm$ 3.0 & N/A \\
 & WPN-CRF & 6.5 $\pm$ 5.0 & 10.3 $\pm$ 5.7 & 10.3 $\pm$ 4.9 & 9.4 $\pm$ 2.7 & N/A \\
% &  L-TapNet+CDT & x & x &  x & x & x \\
 &  Struct NN shot & 15.9 $\pm$ 5.3 & 19.2 $\pm$ 2.9 & 23.1 $\pm$ 2.1 & 26.8 $\pm$ 0.7 & N/A \\
 &  TANL & 32.4 $\pm$ 4.0 & 41.1 $\pm$ 5.0 & 51.7 $\pm$ 2.6 & 58.8 $\pm$ 0.6 & \bf 74.3 $\pm$ 0.2 \\ 
    [\ourspace]
\cmidrule(lr){2-7}
 &  Our model - GloVe & 25.4 $\pm$ 6.1 & 39.7 $\pm$ 2.3 & 52.3 $\pm$ 3.1 & 59.3 $\pm$ 1.4 & 71.8 $\pm$ 0.3 \\
 &  Our model -BERT & \bf 32.7 $\pm$ 3.0 & \bf 43.15 $\pm$ 2.4 & \bf 53.8 $\pm$ 2.7 & \bf 59.8 $\pm$ 1.3 & 71.0 $\pm$ 0.5 \\ 
\end{tabularx}

\begin{tabularx}{\textwidth}{p{\sidedescwidth} p{\modelcolwidth} *{5}{C}}
    \topleftdesc{NCBI-disease}{11} \\
     % & {\bf 1 Shot} & {\bf 5 Shot} & {{\bf 20 Shot}} & {{\bf 50 Shot}} & {{\bf Full Dataset}} \\

\toprule
 & TransferBERT & 16.8 $\pm$ 9.5 & 24.1 $\pm$ 6.3 & 43.0 $\pm$ 5.0 & 56.7 $\pm$ 3.0 & 84.5 $\pm$ 0.9 \\
 & Prototypical Network & 12.2 $\pm$ 8.7 & 12.5 $\pm$ 9.6 & 14.0 $\pm$ 11.6 & 10.8 $\pm$ 7.3 & N/A \\
 & WPN-CRF & 5.5 $\pm$ 4.8 & 6.8 $\pm$ 9.1 & 3.5 $\pm$ 5.4 & 5.7 $\pm$ 5.3 & N/A \\
% &  L-TapNet+CDT & x & x &  x & x & x \\
 &  Struct NN shot & 18.5 $\pm$ 5.6 & 20.6 $\pm$ 5.2 & 27.6 $\pm$ 2.4 & 36.7 $\pm$ 5.0 & N/A \\
 &  TANL & 15.8 $\pm$ 4.0 & 21.0 $\pm$ 6.2 & 26.0 $\pm$ 3.9 & 40.9 $\pm$ 4.2 & 85.8 $\pm$ 0.9 \\ 
    [\ourspace]

  \cmidrule(lr){2-7}
 &  Our model - GloVe & 15.1 $\pm$ 8.7 & 26.2 $\pm$ 6.1 & 44.6 $\pm$ 4.2 & 56.8 $\pm$ 3.1 & \bf 86.7 $\pm$ 0.6 \\
 &  Our model - BERT & \bf 30.7 $\pm$ 9.1 & \bf 34.9 $\pm$ 4.9 & \bf 50.9 $\pm$ 3.3 & \bf 60.5 $\pm$ 2.2 & 85.0 $\pm$ 0.6 \\ 
\end{tabularx}

\begin{tabularx}{\textwidth}{p{\sidedescwidth} p{\modelcolwidth} *{5}{C}}
    \topleftdesc{I2B2-2014}{10} \\
     % & {\bf 1 Shot} & {\bf 5 Shot} & {{\bf 20 Shot}} & {{\bf 50 Shot}} & {{\bf Full Dataset}} \\
\toprule
 & TransferBERT & 58.4 $\pm$ 5.7 & 75.2 $\pm$ 1.9 & 86.2 $\pm$ 0.9 & 90.3 $\pm$ 0.4 & 93.0 $\pm$ 0.1 \\
 & Prototypical Network & 2.1 $\pm$ 0.7 & 2.2 $\pm$ 0.4 & 2.6 $\pm$ 0.4 & 2.7 $\pm$ 0.1 & N/A \\
 & WPN-CRF & 10.0 $\pm$ 2.5 & 13.1 $\pm$ 3.3 & 13.9 $\pm$ 2.1 & 13.3 $\pm$ 2.1 & N/A \\
% &  L-TapNet+CDT & x & x &  x & x & x \\
 &  Struct NN shot & 46.7 $\pm$ 6.4 & 59.1 $\pm$ 1.9 & 67.4 $\pm$ 1.3 & 72.4 $\pm$ 0.6 & N/A \\
 &  TANL & 47.1 $\pm$ 5.2 & 65.1 $\pm$ 2.9 & 80.7 $\pm$  1.2 & 87.0 $\pm$ 0.3 & 92.0 $\pm$ 0.1 \\ 
    [\ourspace]
 
 \cmidrule(lr){2-7}
 &  Our model - GloVe & 58.2 $\pm$ 5.8 & 75.5 $\pm$ 2.3 & 85.6 $\pm$ 1.0 & 90.5 $\pm$ 0.3 & \bf 93.5 $\pm$ 0.1\\
 &  Our model - BERT & \bf 61.9 $\pm$ 4.3 & \bf 76.8 $\pm$ 2.0 & \bf 86.7 $\pm$ 0.8 & \bf 90.5 $\pm$ 0.4 & 93.2 $\pm$ 0.3 \\ 
 \toprule
\end{tabularx}

\caption{Results on held out test sets of all datasets. "Our model - GloVe": this refers to our model with GloVe label encoder. "Our model - BERT": this refers to our model with BERT label encoder. All numbers indicate micro F1 scores unless noted otherwise. Results for low resource settings are average of 10 runs with different support set sampling. Results for high resource setting are average of 5 runs with different random seeds. For some baselines we cannot run the released implementation from originally papers due to GPU out of memory and they are marked as N/A. We visualize the results with bar chart in Appendix \ref{appendix:visualization}.}
% \dr{There are many numbers here; have you tried to see how a bar graph will look? It could be a lot more impressive since it will be easier to see that you almost always win.}
\label{tab:results}

\end{table*}

\section{Experiments}
We evaluate our model and we compare it against existing few shot methods in two scenarios: high resource and low resource (few shot). In both cases, we assume there is a source dataset (which may be a set) with abundant data, and our goal is to maximize model performance on unseen target datasets which follow different taxonomies from the source dataset.

\subsection{Datasets}
We perform experiments on 6 NER datasets from 5 different domains: OntoNotes 5.0 \cite{weischedel2013ontonotes} (Mixed), CoNLL-2003 \cite{tjong-kim-sang-de-meulder-2003-introduction} (News), WNUT-2017 \cite{derczynski-etal-2017-results} (Social), JNLPBA \cite{collier-kim-2004-introduction} (Biology), NCBI-disease \cite{Dogan2014NCBIDC} (Biology) and I2B2-2014 \cite{10.5555/2869975.2870333} (Medical). In all our experiments and following the definition in \ref{model:datasets}, we treat OntoNotes as the \textbf{source dataset} and all other as \textbf{target datasets}.\footnote{We use train/dev/test split released from CoNLL-2012 shared task: \url{https://cemantix.org/conll/2012/data.html}.}

\subsection{Settings and Evaluation}
% \mb{Jie: refer to the notation section where we present the concept of SOURCE and TARGET, and explain what is source and what is target for us}
In this Section, we present the different experiments, and how do we carry out the evaluation.

\paragraph{High Resource:} Given a target dataset, we simply take all available data and evaluate on the standard held-out test set.

\paragraph{Low Resource:} Given a target dataset, we downsample the data (at sentence level) in the train split to construct a $K$-shot support set. This simulates the low resource scenario where only a few training examples are available in the target dataset. 
The definition of a $K$-shot support set is that it contains exact $K$ examples for each of the labels. However, unlike the text classification task where each sentence is associated with one label, in the NER task multiple named entities may co-occur in the same sentence. We cannot guarantee that the support set contains exact $K$ named entities for each label after downsampling. We therefore define the proxy for $K$-shot support set similar as the one by \newcite{hou-etal-2020-shot}, with the following two criteria: 1) Each label in the target dataset (except ``O'') has at least $K$ corresponding named entities in the support set; 2) At least one of the labels in the target dataset will have less than $K$ named entities in the support set if any sentence is removed.\footnote{We count at named entity level instead of token level. For example, ``Lionel Messi'' is counted as one occurrence for PERSON entity. However, \newcite{hou-etal-2020-shot} counted it as one occurrence for ``B-PERSON'' (for token ``Lionel'') and one occurrence for ``I-PERSON'' (for token ``Messi'').} We apply the same downsampling algorithm as in \cite{hou-etal-2020-shot} for the support set. More details can be found in Appendix \ref{apppendix:sampling_algorithm}.

To evaluate the model performance in the $K$-shot support set, most prior work \cite{hou-etal-2020-shot, athiwaratkun2020augmented, Fritzler_2019} followed the few-shot classification setup, where test sets are also downsampled to $K$-shot subsets (query set) such that each entity labels are evenly distributed. The model is trained and evaluated on multiple support datasets and query set pairs, and final model performance is reported with average of scores on each query set. However, we argue that in real world cases, entity labels have certain distribution corresponding to the domain, downsampled $K$-shot query set does not reflect this real distribution. Therefore instead of evaluating on the downsampled query set, we directly evaluate the model in the full test split from the target dataset. This also improves comparability and replicability of our results since the same test set is used across and in prior work (even in papers that are not focused on few-shot experiments).

\paragraph{Evaluation} To thoroughly test our model, we evaluate it with 1-shot, 5-shot, 20-shot, 50-shot (low resource) and also the full dataset (high resource) settings. Following prior work \cite{conll2003}, we use micro F1 score as metric. For low resource settings, we repeat the experiments 10 times with randomly sampled support sets. For high resource setting, we repeat the experiments 5 times with different random seeds. In all cases, we report average micro F1 with standard deviation. Table \ref{tab:stats-table} shows an overview of dataset statistics.

\subsection{Baselines}
% \mb{Jie: it is not clear if these models are DIRECTLY comparable to ours. We only specify major differences with TANL, because of the nature and the pretrained model, but how about the rest of the model. make sure to specify which ones are directly comparable and which ones are not.}
\textbf{TransferBERT} trains the same NER model in \cite{devlin2019bert} by pre-finetuning on a source dataset then finetuning on a target dataset. \textbf{Prototypical Network} \cite{snell2017prototypical} approaches NER as a token level classification task. It assigns label for each token based on similarities between candidate token and tokens in few shot support set. \textbf{WPN-CRF} \cite{Fritzler_2019} pretrains a prototypical network with source dataset and evaluate it on target dataset without finetuning. It uses a conditional random field (CRF) \cite{huang2015bidirectional} to output the final labels of the sentence. \textbf{Struct NN shot} \cite{yang2020simple} finds nearest token in support set for a given candidate token and assign it the same label as its nearest neighbor. \textbf{TANL} \cite{paolini2021structured} forms NER as sequence to sequence. The model is trained to generate the original input text with entities being decorated in a bracket.\footnote{We are not able to include \cite{hou-etal-2020-shot} as a baseline as we are not able to reproduce the model with their published repository, even on a machine with 40GB of GPU memory. We also cannot compare with the published results due to the differences in the following settings: (1) we are evaluating our model on full test splits while Hou et al. carry out an episodic evaluation) and (2) We use more datasets (from different domains).}%\footnote{An example for TANL: given an input sentence ``Adam Smith worked for University of Edinburgh'', the output of TANL model will be ``[ Adam Smith | person ] worked for [ University of Edinburgh | organization ]''.}

\subsection{Hyperparameters}
We use English cased BERT-base \cite{devlin2019bert} as contextual embedder for all baseline models and our model, except for TANL where T5-base is used.\footnote{We use the checkpoint released for BERT-base: \url{https://github.com/google-research/bert}, and checkpoints released in Hugging Face for T5-base: \url{https://huggingface.co/t5-base}} We use Adam optimizer \cite{kingma2014adam} to train our model with a learning rate of $\num{1e-5}$ and batch size of 10. We pre-finetune our model on the source dataset (Ontonotes) for 3 epochs and continue finetuning on target datasets for 200 epochs for both high resource and low resource settings. We pick the last epoch as the final model. For label names, we manually expand all shortcut names into full natural language names (e.g. ``PER'' to ``person'', ``LOC'' to ``location'') and lower case all names. Textual forms for all datasets can be found in Appendix \ref{appendix:datasets_label_names}. We run all experiments on NVIDIA V100 GPU.\footnote{More details about hardware in Appendix \ref{appendix:hardware}.}

\subsection{GloVe as Label Encoder}
\label{experiment:glove-as-label-encoder}
We experiment with GloVe embeddings \cite{pennington2014glove} as the label encoder.\footnote{We use 300 dimensional GloVe that is pretrained on Wikipedia and Gigaword 5 corpus released here: \url{https://nlp.stanford.edu/projects/glove/}} In this case, our model has no extra parameters compared to other baselines. As in the case with BERT, the vectors are updated throughout the training. Given that there is no [CLS] token available, we apply max pooling on all the GloVe embeddings corresponding to each label token. If the label consists only of one token, max pooling will return the actual GloVe embedding for the token as the label representation.
\begin{table}[H]
%\small
\begin{center}
\begin{tabular}{lccccc}
%\hline
\toprule
\multirow{2}{*}{\bf Dataset} & \multicolumn{4}{c}{\bf Support Set Shot} \\
\cline{2-5}
 & \bf 1 & \bf 5 & \bf 20 & \bf 50  \\
\toprule
CoNLL'03 & 3.6 & 12.3 & 38.5 & 102.5 \\
WNUT'17 & 13.4 & 44.6 & 143.6 & 366.3 \\
JNLPBA & 6.8 & 27.5 & 99.2 & 241.2 \\
NCBI & 1.8 & 3.7 & 14.5 & 37.2 \\
I2B2'14 & 155.4 & 613.4 & 2339.4 & 5888.1 \\
\toprule
\end{tabular}
\end{center}
\caption{\label{tab:stats-table} Number of sentences in support set with different shots for all target datasets. Numbers are averaged across 10 different random samplings. NCBI refers to NCBI-disease dataset. More details are reported in Appendix \ref{appendix:datasets_stats}.}
\end{table}

\subsection{Results} \label{exp:results}
We summarize experiment results in Table \ref{tab:results}. As shown, our model outperforms all previous methods in low resource settings. In extreme low resource scenarios (1 and 5 shot), our model performs significantly better than previous methods by a margin of 6.6 F1 and 4.8 F1 on average in 1 shot and 5 shot, respectively. This indicates that our model can leverage semantics in label names effectively to improve accuracy when data is extremely scarce. However, we also notice that when the target data size increases, the improvement of our model becomes smaller. This suggests that with more training examples, the model relies less on semantics of labels.

In a high resource setting,  we find that our model achieves the same level of performance as other baselines, except for JNLPBA dataset where our model is 3.3 F1 behind TANL.\footnote{We cannot run released implementation of three baselines (marked as N/A in Table \ref{tab:results}) due to GPU out of memory even with 40GB of GPU memory.} This model is based on T5-base which is pretrained on a much larger unannotated dataset, and with different objectives, than our BERT-base encoders.
% Also, we need to point out that our model has significant advantages against TANL on training time and latency (see detailed analysis on XX). \jm{Add section number}

We also note that when label names in the target dataset are similar to the source ones, few shot models have a much smaller gap with their high resource counterparts, compared to when source and target label names are totally different. Specifically, CoNLL-2003, WNUT-2017 and I2B2 have more similar label names with Ontonotes (the source data), and our model can achieve 84\%, 91\% and 83\% of the score of the high resource model performance with only 5 shot. While for JNLPBA and NCBI-disease, where the label names are totally different from source data, our model can only achieve 61\% and 41\% of the score of the high resource model performance with 5 shot.

\section{Analysis}
\label{exp:analysis}
Here, we show how semantics in label names help in low resource scenarios and how our model benefits from pre-finetuning stage.

\begin{table}[H]
\begin{center}
\begin{tabular}{lccc}
%\hline
\toprule
\multirow{2}{*}{\makecell{\bf Entity \\ \bf Types}} & \multicolumn{2}{c}{\bf Original Labels} & \bf Renamed Labels \\
\cline{2-3} \cline{4-4} 
 & \bf 0 shot & \bf 1 shot & \bf 0 shot \\
\toprule
PER & 92.3 & 90.3 & 85.4 \\
LOC & 70.9 & 61.2 & 54.8 \\
ORG & 50.3 & 59.7 & 58.4 \\
MISC & 0.5 & 47.5 & 6.8 \\
\toprule
\end{tabular}
\end{center}
\caption{\label{tab:zero-shot} F1 for 0 and 1 shot performance on CoNLL-2003 development set.}
\end{table}

\subsection{Impact of the Label Encoder}

We hypothesize that encoding label names with a label encoder (either BERT or GloVe) leverages prior knowledge from the pretraining phase and uses it as inductive bias. In addition, by performing pre-finetuning on the source dataset, we are not only aligning the embedding space between labels and tokens in the vocabulary, but also updating the label encoder to produce useful label representations in the source dataset. 

To further strengthen our hypothesis (besides what is presented in Table \ref{tab:results}), we show results in zero shot settings. Specifically, we pre-finetune a model on the source dataset (Ontonotes) and directly test it on CoNLL-2003 without updating its parameters. We also rename the labels to avoid overlapping of label names between source and target datasets while still retaining the semantics.\footnote{Ontonotes has both ``LOC'' and ``GPE'' labels, however, the definition of label ``GPE'' in Ontonotes is much closer to ``LOC'' in CoNLL2003. Therefore, we use "GPE" instead of ``LOC'' for zero shot experiments.} Particularly, during evaluation we rename ``PER'' to ``individual'', ``LOC'' to ``geographical area'' and ``ORG'' to ``corporation''. ``MISC'' stays the same since it does not overlap with any of the Ontonotes labels. The results are shown in Table \ref{tab:zero-shot}. 

With original label names, the zero shot performance of our model is comparable to 1 shot performance for all entity types with the exception of ``MISC''. Even with the renamed labels that do not have any overlap with the source dataset, the zero shot performance still remains comparable with 1 shot. This seems to validate our hypothesis that the model is able to leverage prior knowledge.

% Experiment where we change label name at inference time. We can also explore how similar labels are cluster together with each other, also how labels and entity values are clustered together. This will need to compare with baseline models.

\subsection{Semantics of Label Names}
To demonstrate the impact of semantics of label names, we carry out experiments with our model on target datasets with the following variations of label names: (1) original label names (which is simply our experimental setup as in the experiments above, where we use the natural language form of the label names), (2) meaningless label names and (3) misleading label names. 

We compare our model with the TransferBERT baseline, since it is the counterpart of our model without label semantics. We pre-finetune our model on Ontonotes as previous experiments. Results on CoNLL2003 and JNLPBA are shown in Figure \ref{fig:results_label_words}.\footnote{We present experiments with contextualized label representations in appendix \ref{appendix:contextual_label_rep}.}

\paragraph{Meaningless labels}
We simply use ``label 1'', ``label 2'' etc., as input representation for label names, which simulates the case where there is no more semantics information in the form than the fact that they are different labels and they have some sort of ordering. This evaluates the few shot model performance when meaningless (or shallow in semantics, just a differentiation of label indices) inputs are given. Comparing to the original label names, the results drop in 1 and 5 shot settings, then gradually converged to the original label performance as the training data size increases. This shows that label semantics is critical for extreme low resource scenarios (1 and 5 shot). 

\begin{center}
  \includegraphics[scale=0.46]{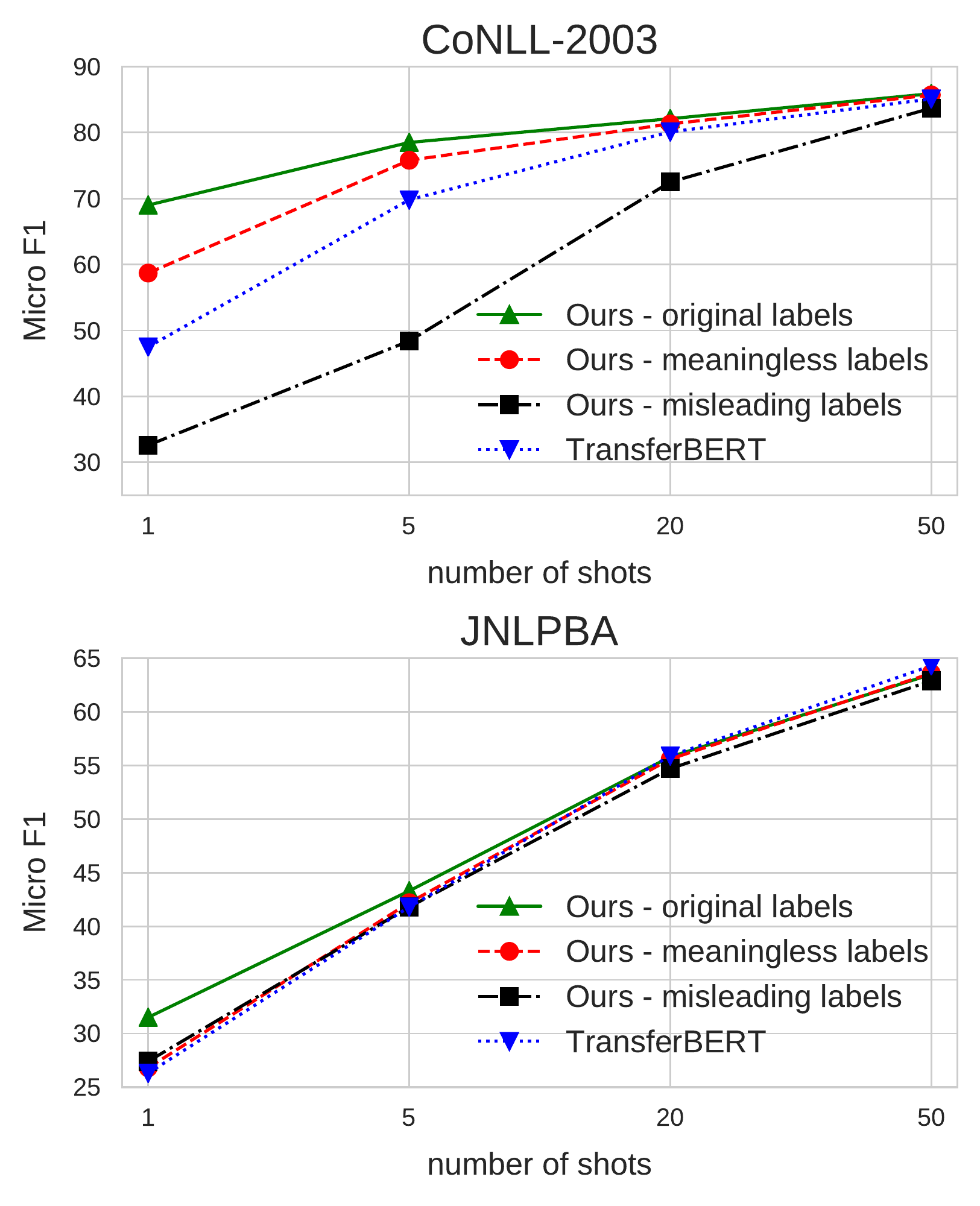}
  \captionof{figure}{Model performance on meaningless and misleading laberls. Micro F1 is reported on the development data.}
  \label{fig:results_label_words}
\end{center}

\paragraph{Misleading labels} 

We randomly  swap the natural language form between labels. For example, in CoNLL2003 dataset, we assign ``location'' for ``PER'', ``person'' for ``ORG'', ``organization'' for ``MISC'' and ``miscellaneous'' for ``PER''.\footnote{For each run we randomly assign different misleading label names, and we report results averaging 10 different runs.} The performance drops are larger for CoNLL2003 than the ones in JNLPBA. We hypothesize that since CoNLL2003 label set is closer to Ontonotes, there is stronger prior knowledge incorporated in the label encoder from the pre-finetuning phase. Also, we find that more supervised examples are required to correct such wrong strong prior information. JNLPBA needs 5 shot data to achieve the same performance with original labels and misleading labels, but  CoNLL2003 needs 50 shot data to match the performance. This indicates that our model is misled by the labels when the number of training examples is small, which indicates that the label semantics signal is critical in few shot settings.  %It indicates that our model is misled by the labels when the actual data in examples is too low. With enough data, it can memorize and override the label semantics signal by the number of examples.

\subsection{Impact of Pre-finetuning}

Our model does not require a new randomly initialized top layer classifier for a new dataset, we hypothesize that it can prevent the model from forgetting learned prior knowledge from the pre-finetuning stage thus benefits the low resource scenarios, where prior knowledge is critical. To validate it, we compare 1-shot results on target datasets with and without pre-finetuning stage, as shown in Table \ref{tab:no-pre-finetune}. 
First, when pre-finetuning stage is eliminated, performance of both our model and TransferBERT drop significantly, indicating that prior knowledge from pre-finetuning stage is critical in low resource settings. Second, our model outperforms TransferBERT significantly when pre-finetuning stage is included, however, the performance is similar between our model and TransferBERT when it is excluded. This suggests that our model is highly effective in leveraging knowledge learned from the pre-finetuning stage. 

\begin{table}[!ht]
\begin{center}
\small
\begin{tabular}{lcccc}
%\hline
\toprule
\multirow{2}{*}{\makecell{\bf Datasets}} & \multicolumn{2}{c}{\makecell{\bf Pre-finetune on \\ \bf Ontonotes}} & \multicolumn{2}{c}{\makecell{\bf No \\ \bf pre-finetune}} \\
\cline{2-3} \cline{4-5} 
 & \makecell{\bf Transfer- \\ \bf BERT} & \bf Ours & \makecell{\bf Transfer- \\ \bf BERT} & \bf Ours \\
\toprule
CoNLL'03 & 47.5 & 69.0 & 9.0 & 10.7\\
WNUT'17 & 35.6 & 48.2 & 4.0 & 5.7\\
JNLPBA & 26.3 & 31.5 & 14.8 & 19.5\\
NCBI & 15.1 & 31.3 & 12.5 & 13.9 \\
I2B2'14 & 56.9 & 60.1 & 47.5 & 46.8 \\
\toprule
\end{tabular}
\end{center}
\caption{\label{tab:no-pre-finetune} 1-shot performance on development set of corresponding datasets. Micro F1 is reported. NCBI refers to NCBI-disease dataset.}
\end{table}

\section{Related Work}

%\mb{Jie: do we need this whole paragraph? I think we mention this in the analysis} Comparing to similarity-based methods, our model not only leverages semantics in label names but also avoids representing labels by averaging embedding of examples in training data, which makes it suitable for both low and high resource settings.
% Also, our model works in the discriminative fashion, thus it will not suffer from large decoding latency as the aforementioned sequence to sequence methods \cite{athiwaratkun2020augmented, paolini2021structured, DBLP:journals/corr/abs-2005-14165}.
%In addition, most NER models randomly initialize the output layer decoder for new labels during finetuning; our model does not have this limitation since it leverages the encoder priors (it can be pretrained with masked language modeling objectives \cite{devlin2019bert} in a shared vocabulary with words). Both the word and label encoder can be reused for any new labels which allows for knowledge transferring from existing domains to a new domain without re-initializing the top classifier.

% \dr{Since we are positioning it from the perspective of Label Semantics, we need to go all the way back to Chang et al. AAAI'08, \url{http://cogcomp.seas.upenn.edu/papers/CRRS08.pdf}  Also, there are some recent papers that mention label/task understanding, like the PET papers from ‪Hinrich Schütze (with Timo Schick). Need to check for additional recent papers in ACL/NAACL this year.}
\paragraph{Few Shot Learning:} Meta learning is widely studied for the problem of few shot learning, aiming to quickly adapt a model to new tasks based on tasks learned in an earlier stage. Recent research \cite{snell2017prototypical, vinyals2017matching, DBLP:journals/corr/abs-1711-06025} mostly focused on metric-based methods. \citet{snell2017prototypical} learns a prototype representation for each class and classify test data based on their similarities with prototypes. These methods have been successfully adapted to NLP tasks such as classification \cite{DBLP:journals/corr/abs-1805-07513, DBLP:journals/corr/abs-1908-06039}, relation classification \cite{DBLP:journals/corr/abs-1810-10147} and NER \cite{Fritzler_2019, yang2020simple}. However, all these methods do not directly leverage the semantics of label names. % Also, once the models are trained on source datasets, they are applied in target datasets without further finetuning, which under utilize the target domain training data especially in high resource setting. \dr{I was not sure what's the importance of the last sentence; if you think that it is important, clarify that you are referring to earlier models.}

\paragraph{Label Semantics:} Earlier work %\cite{Chang2008ImportanceOS, 7472838, Luo2021DontMT} 
has shown the ability to perform zero- and few-shot learning by exploiting the semantic of labels in text classification tasks
~\cite{Chang2008ImportanceOS, Luo2021DontMT}. %is less invesigated. 
\citet{Zhou2018ZeroShotOE} study zero-shot fine-type NER with label semantics by automatically reading from Wikipedia via a linking approach, but assumes that the mentions of the entities are given. \citet{paolini2021structured} and \citet{athiwaratkun2020augmented} approach NER as a generation task and predict named entities in augmented (or decorated) languages. \citet{DBLP:journals/corr/abs-2106-01760} reformulate NER as a cloze task and use sequence to sequence models to fill named entities in pre-defined templates. Both of these two methods suffer from long inference time due to an autoregressive decoder. \citet{hou-etal-2020-shot} leverage label semantics in Task-Adaptive Projection Network (TapNet), where the core idea is to learn a projection function that separates words that have different labels in the projected space. In contrast, our model learns to align token representations with label representations. \citet{hou-etal-2020-shot} only uses label representations as a reference to guide the learning of the projection function, and in their case label representations are computed once. Our label representations are updated with every update while training.
% Different from our model, where label representations are learned end to end to align with token representations, \citet{hou-etal-2020-shot} only uses label representations as references to guide the learning of the projection function, and label representations are computed once and are then fixed.

\section{Conclusion}
We propose a neural architecture that leverages semantics of label names for Named Entity Recognition. Our model significantly outperforms the state-of-the-art few shot NER baselines on low resource settings, and performs on-par in the high resource setting. We perform extensive experiments to show that the label encoder incorporates strong prior knowledge from BERT and a dataset (source dataset) used in a pre-finetuning stage. We demonstrate that the semantics of label names in target datasets are critical to retrieve the prior knowledge. We also show that our model is robust to variation of label names and that it is able to differentiate between semantically closed labels.

% \bibliography{tacl2018}
% \bibliographystyle{acl_natbib}

\bibliography{anthology,custom}

\begin{thebibliography}{43}
\expandafter\ifx\csname natexlab\endcsname\relax\def\natexlab#1{#1}\fi

\bibitem[{Aghajanyan et~al.(2021)Aghajanyan, Gupta, Shrivastava, Chen,
  Zettlemoyer, and Gupta}]{DBLP:journals/corr/abs-2101-11038}
Armen Aghajanyan, Anchit Gupta, Akshat Shrivastava, Xilun Chen, Luke
  Zettlemoyer, and Sonal Gupta. 2021.
\newblock \href {http://arxiv.org/abs/2101.11038} {Muppet: Massive multi-task
  representations with pre-finetuning}.
\newblock \emph{CoRR}, abs/2101.11038.

\bibitem[{Athiwaratkun et~al.(2020)Athiwaratkun, dos Santos, Krone, and
  Xiang}]{athiwaratkun2020augmented}
Ben Athiwaratkun, Cicero~Nogueira dos Santos, Jason Krone, and Bing Xiang.
  2020.
\newblock \href {http://arxiv.org/abs/2009.13272} {Augmented natural language
  for generative sequence labeling}.

\bibitem[{Bao et~al.(2019)Bao, Wu, Chang, and
  Barzilay}]{DBLP:journals/corr/abs-1908-06039}
Yujia Bao, Menghua Wu, Shiyu Chang, and Regina Barzilay. 2019.
\newblock \href {http://arxiv.org/abs/1908.06039} {Few-shot text classification
  with distributional signatures}.
\newblock \emph{CoRR}, abs/1908.06039.

\bibitem[{Carreras et~al.(2003)Carreras, M{\`a}rquez, and
  Padr{\'o}}]{carreras-etal-2003-learning}
Xavier Carreras, Llu{\'\i}s M{\`a}rquez, and Llu{\'\i}s Padr{\'o}. 2003.
\newblock \href {https://aclanthology.org/W03-0422} {Learning a
  perceptron-based named entity chunker via online recognition feedback}.
\newblock In \emph{Proceedings of the Seventh Conference on Natural Language
  Learning at {HLT}-{NAACL} 2003}, pages 156--159.

\bibitem[{Chan and Roth(2011)}]{chan-roth-2011-exploiting}
Yee~Seng Chan and Dan Roth. 2011.
\newblock \href {https://aclanthology.org/P11-1056} {Exploiting
  syntactico-semantic structures for relation extraction}.
\newblock In \emph{Proceedings of the 49th Annual Meeting of the Association
  for Computational Linguistics: Human Language Technologies}, pages 551--560,
  Portland, Oregon, USA. Association for Computational Linguistics.

\bibitem[{Chang et~al.(2008)Chang, Ratinov, Roth, and
  Srikumar}]{Chang2008ImportanceOS}
Ming-Wei Chang, Lev-Arie Ratinov, Dan Roth, and Vivek Srikumar. 2008.
\newblock Importance of semantic representation: Dataless classification.
\newblock In \emph{AAAI}.

\bibitem[{Chen et~al.(2020)Chen, Wang, Tian, Yang, and
  Yang}]{DBLP:journals/corr/abs-2010-01677}
Jiaao Chen, Zhenghui Wang, Ran Tian, Zichao Yang, and Diyi Yang. 2020.
\newblock \href {http://arxiv.org/abs/2010.01677} {Local additivity based data
  augmentation for semi-supervised {NER}}.
\newblock \emph{CoRR}, abs/2010.01677.

\bibitem[{Clark and Manning(2015)}]{clark-manning-2015-entity}
Kevin Clark and Christopher~D. Manning. 2015.
\newblock \href {https://doi.org/10.3115/v1/P15-1136} {Entity-centric
  coreference resolution with model stacking}.
\newblock In \emph{Proceedings of the 53rd Annual Meeting of the Association
  for Computational Linguistics and the 7th International Joint Conference on
  Natural Language Processing (Volume 1: Long Papers)}, pages 1405--1415,
  Beijing, China. Association for Computational Linguistics.

\bibitem[{Collier and Kim(2004)}]{collier-kim-2004-introduction}
Nigel Collier and Jin-Dong Kim. 2004.
\newblock \href {https://aclanthology.org/W04-1213} {Introduction to the
  bio-entity recognition task at {JNLPBA}}.
\newblock In \emph{Proceedings of the International Joint Workshop on Natural
  Language Processing in Biomedicine and its Applications
  ({NLPBA}/{B}io{NLP})}, pages 73--78, Geneva, Switzerland. COLING.

\bibitem[{Collobert et~al.(2011)Collobert, Weston, Bottou, Karlen, Kavukcuoglu,
  and Kuksa}]{DBLP:journals/corr/abs-1103-0398}
Ronan Collobert, Jason Weston, L{\'{e}}on Bottou, Michael Karlen, Koray
  Kavukcuoglu, and Pavel~P. Kuksa. 2011.
\newblock \href {http://arxiv.org/abs/1103.0398} {Natural language processing
  (almost) from scratch}.
\newblock \emph{CoRR}, abs/1103.0398.

\bibitem[{Cui et~al.(2021)Cui, Wu, Liu, Yang, and
  Zhang}]{DBLP:journals/corr/abs-2106-01760}
Leyang Cui, Yu~Wu, Jian Liu, Sen Yang, and Yue Zhang. 2021.
\newblock \href {http://arxiv.org/abs/2106.01760} {Template-based named entity
  recognition using {BART}}.
\newblock \emph{CoRR}, abs/2106.01760.

\bibitem[{Derczynski et~al.(2017)Derczynski, Nichols, van Erp, and
  Limsopatham}]{derczynski-etal-2017-results}
Leon Derczynski, Eric Nichols, Marieke van Erp, and Nut Limsopatham. 2017.
\newblock \href {https://doi.org/10.18653/v1/W17-4418} {Results of the
  {WNUT}2017 shared task on novel and emerging entity recognition}.
\newblock In \emph{Proceedings of the 3rd Workshop on Noisy User-generated
  Text}, pages 140--147, Copenhagen, Denmark. Association for Computational
  Linguistics.

\bibitem[{Devlin et~al.(2019)Devlin, Chang, Lee, and
  Toutanova}]{devlin2019bert}
Jacob Devlin, Ming-Wei Chang, Kenton Lee, and Kristina Toutanova. 2019.
\newblock \href {http://arxiv.org/abs/1810.04805} {Bert: Pre-training of deep
  bidirectional transformers for language understanding}.

\bibitem[{Ding et~al.(2021)Ding, Xu, Chen, Wang, Han, Xie, Zheng, and
  Liu}]{DBLP:journals/corr/abs-2105-07464}
Ning Ding, Guangwei Xu, Yulin Chen, Xiaobin Wang, Xu~Han, Pengjun Xie,
  Hai{-}Tao Zheng, and Zhiyuan Liu. 2021.
\newblock \href {http://arxiv.org/abs/2105.07464} {Few-nerd: {A} few-shot named
  entity recognition dataset}.
\newblock \emph{CoRR}, abs/2105.07464.

\bibitem[{Dogan et~al.(2014)Dogan, Leaman, and Lu}]{Dogan2014NCBIDC}
Rezarta~Islamaj Dogan, Robert Leaman, and Zhiyong Lu. 2014.
\newblock Ncbi disease corpus: A resource for disease name recognition and
  concept normalization.
\newblock \emph{Journal of biomedical informatics}, 47:1--10.

\bibitem[{Finn et~al.(2017)Finn, Abbeel, and Levine}]{finn2017modelagnostic}
Chelsea Finn, Pieter Abbeel, and Sergey Levine. 2017.
\newblock \href {http://arxiv.org/abs/1703.03400} {Model-agnostic meta-learning
  for fast adaptation of deep networks}.

\bibitem[{Fritzler et~al.(2019)Fritzler, Logacheva, and Kretov}]{Fritzler_2019}
Alexander Fritzler, Varvara Logacheva, and Maksim Kretov. 2019.
\newblock \href {https://doi.org/10.1145/3297280.3297378} {Few-shot
  classification in named entity recognition task}.
\newblock \emph{Proceedings of the 34th ACM/SIGAPP Symposium on Applied
  Computing}.

\bibitem[{Han et~al.(2018)Han, Zhu, Yu, Wang, Yao, Liu, and
  Sun}]{DBLP:journals/corr/abs-1810-10147}
Xu~Han, Hao Zhu, Pengfei Yu, Ziyun Wang, Yuan Yao, Zhiyuan Liu, and Maosong
  Sun. 2018.
\newblock \href {http://arxiv.org/abs/1810.10147} {Fewrel: {A} large-scale
  supervised few-shot relation classification dataset with state-of-the-art
  evaluation}.
\newblock \emph{CoRR}, abs/1810.10147.

\bibitem[{Hou et~al.(2020)Hou, Che, Lai, Zhou, Liu, Liu, and
  Liu}]{hou-etal-2020-shot}
Yutai Hou, Wanxiang Che, Yongkui Lai, Zhihan Zhou, Yijia Liu, Han Liu, and Ting
  Liu. 2020.
\newblock \href {https://doi.org/10.18653/v1/2020.acl-main.128} {Few-shot slot
  tagging with collapsed dependency transfer and label-enhanced task-adaptive
  projection network}.
\newblock In \emph{Proceedings of the 58th Annual Meeting of the Association
  for Computational Linguistics}, pages 1381--1393, Online. Association for
  Computational Linguistics.

\bibitem[{Huang et~al.(2020)Huang, Li, Subudhi, Jose, Balakrishnan, Chen, Peng,
  Gao, and Han}]{DBLP:journals/corr/abs-2012-14978}
Jiaxin Huang, Chunyuan Li, Krishan Subudhi, Damien Jose, Shobana Balakrishnan,
  Weizhu Chen, Baolin Peng, Jianfeng Gao, and Jiawei Han. 2020.
\newblock \href {http://arxiv.org/abs/2012.14978} {Few-shot named entity
  recognition: {A} comprehensive study}.
\newblock \emph{CoRR}, abs/2012.14978.

\bibitem[{Huang et~al.(2015)Huang, Xu, and Yu}]{huang2015bidirectional}
Zhiheng Huang, Wei Xu, and Kai Yu. 2015.
\newblock \href {http://arxiv.org/abs/1508.01991} {Bidirectional lstm-crf
  models for sequence tagging}.

\bibitem[{Karpukhin et~al.(2020)Karpukhin, Oguz, Min, Wu, Edunov, Chen, and
  Yih}]{DBLP:journals/corr/abs-2004-04906}
Vladimir Karpukhin, Barlas Oguz, Sewon Min, Ledell Wu, Sergey Edunov, Danqi
  Chen, and Wen{-}tau Yih. 2020.
\newblock \href {http://arxiv.org/abs/2004.04906} {Dense passage retrieval for
  open-domain question answering}.
\newblock \emph{CoRR}, abs/2004.04906.

\bibitem[{Kingma and Ba(2014)}]{kingma2014adam}
Diederik~P Kingma and Jimmy Ba. 2014.
\newblock Adam: A method for stochastic optimization.
\newblock \emph{arXiv preprint arXiv:1412.6980}.

\bibitem[{Lample et~al.(2016)Lample, Ballesteros, Subramanian, Kawakami, and
  Dyer}]{lample-etal-2016-neural}
Guillaume Lample, Miguel Ballesteros, Sandeep Subramanian, Kazuya Kawakami, and
  Chris Dyer. 2016.
\newblock \href {https://doi.org/10.18653/v1/N16-1030} {Neural architectures
  for named entity recognition}.
\newblock In \emph{Proceedings of the 2016 Conference of the North {A}merican
  Chapter of the Association for Computational Linguistics: Human Language
  Technologies}, pages 260--270, San Diego, California. Association for
  Computational Linguistics.

\bibitem[{Lison et~al.(2020)Lison, Barnes, Hubin, and
  Touileb}]{lison-etal-2020-named}
Pierre Lison, Jeremy Barnes, Aliaksandr Hubin, and Samia Touileb. 2020.
\newblock \href {https://doi.org/10.18653/v1/2020.acl-main.139} {Named entity
  recognition without labelled data: A weak supervision approach}.
\newblock In \emph{Proceedings of the 58th Annual Meeting of the Association
  for Computational Linguistics}, pages 1518--1533, Online. Association for
  Computational Linguistics.

\bibitem[{Logeswaran et~al.(2019)Logeswaran, Chang, Lee, Toutanova, Devlin, and
  Lee}]{logeswaran-etal-2019-zero}
Lajanugen Logeswaran, Ming-Wei Chang, Kenton Lee, Kristina Toutanova, Jacob
  Devlin, and Honglak Lee. 2019.
\newblock \href {https://doi.org/10.18653/v1/P19-1335} {Zero-shot entity
  linking by reading entity descriptions}.
\newblock In \emph{Proceedings of the 57th Annual Meeting of the Association
  for Computational Linguistics}, pages 3449--3460, Florence, Italy.
  Association for Computational Linguistics.

\bibitem[{Luo et~al.(2021)Luo, Liu, Lin, and Zhang}]{Luo2021DontMT}
Qiaoyang Luo, Lingqiao Liu, Yuhao Lin, and Wei Zhang. 2021.
\newblock Don’t miss the labels: Label-semantic augmented meta-learner for
  few-shot text classification.
\newblock In \emph{FINDINGS}.

\bibitem[{Moll{\'a} et~al.(2006)Moll{\'a}, van Zaanen, and
  Smith}]{molla-etal-2006-named}
Diego Moll{\'a}, Menno van Zaanen, and Daniel Smith. 2006.
\newblock \href {https://aclanthology.org/U06-1009} {Named entity recognition
  for question answering}.
\newblock In \emph{Proceedings of the Australasian Language Technology Workshop
  2006}, pages 51--58, Sydney, Australia.

\bibitem[{Paolini et~al.(2021)Paolini, Athiwaratkun, Krone, Ma, Achille,
  Anubhai, dos Santos, Xiang, and Soatto}]{paolini2021structured}
Giovanni Paolini, Ben Athiwaratkun, Jason Krone, Jie Ma, Alessandro Achille,
  Rishita Anubhai, Cicero~Nogueira dos Santos, Bing Xiang, and Stefano Soatto.
  2021.
\newblock \href {http://arxiv.org/abs/2101.05779} {Structured prediction as
  translation between augmented natural languages}.

\bibitem[{Pennington et~al.(2014)Pennington, Socher, and
  Manning}]{pennington2014glove}
Jeffrey Pennington, Richard Socher, and Christopher~D. Manning. 2014.
\newblock \href {http://www.aclweb.org/anthology/D14-1162} {Glove: Global
  vectors for word representation}.
\newblock In \emph{Empirical Methods in Natural Language Processing (EMNLP)},
  pages 1532--1543.

\bibitem[{Snell et~al.(2017)Snell, Swersky, and Zemel}]{snell2017prototypical}
Jake Snell, Kevin Swersky, and Richard~S. Zemel. 2017.
\newblock \href {http://arxiv.org/abs/1703.05175} {Prototypical networks for
  few-shot learning}.

\bibitem[{Stubbs and Uzuner(2015)}]{10.5555/2869975.2870333}
Amber Stubbs and \"{O}zlem Uzuner. 2015.
\newblock Annotating longitudinal clinical narratives for de-identification.
\newblock \emph{J. of Biomedical Informatics}, 58(S):S20–S29.

\bibitem[{Sung et~al.(2017)Sung, Yang, Zhang, Xiang, Torr, and
  Hospedales}]{DBLP:journals/corr/abs-1711-06025}
Flood Sung, Yongxin Yang, Li~Zhang, Tao Xiang, Philip H.~S. Torr, and
  Timothy~M. Hospedales. 2017.
\newblock \href {http://arxiv.org/abs/1711.06025} {Learning to compare:
  Relation network for few-shot learning}.
\newblock \emph{CoRR}, abs/1711.06025.

\bibitem[{Tjong Kim~Sang and
  De~Meulder(2003{\natexlab{a}})}]{tjong-kim-sang-de-meulder-2003-introduction}
Erik~F. Tjong Kim~Sang and Fien De~Meulder. 2003{\natexlab{a}}.
\newblock \href {https://aclanthology.org/W03-0419} {Introduction to the
  {C}o{NLL}-2003 shared task: Language-independent named entity recognition}.
\newblock In \emph{Proceedings of the Seventh Conference on Natural Language
  Learning at {HLT}-{NAACL} 2003}, pages 142--147.

\bibitem[{Tjong Kim~Sang and De~Meulder(2003{\natexlab{b}})}]{conll2003}
Erik~F. Tjong Kim~Sang and Fien De~Meulder. 2003{\natexlab{b}}.
\newblock \href {https://doi.org/10.3115/1119176.1119195} {Introduction to the
  conll-2003 shared task: Language-independent named entity recognition}.
\newblock In \emph{Proceedings of the Seventh Conference on Natural Language
  Learning at HLT-NAACL 2003 - Volume 4}, CONLL '03, page 142–147, USA.
  Association for Computational Linguistics.

\bibitem[{Vinyals et~al.(2017)Vinyals, Blundell, Lillicrap, Kavukcuoglu, and
  Wierstra}]{vinyals2017matching}
Oriol Vinyals, Charles Blundell, Timothy Lillicrap, Koray Kavukcuoglu, and Daan
  Wierstra. 2017.
\newblock \href {http://arxiv.org/abs/1606.04080} {Matching networks for one
  shot learning}.

\bibitem[{Vyas and Ballesteros(2020)}]{DBLP:journals/corr/abs-2010-11333}
Yogarshi Vyas and Miguel Ballesteros. 2020.
\newblock \href {http://arxiv.org/abs/2010.11333} {Linking entities to unseen
  knowledge bases with arbitrary schemas}.
\newblock \emph{CoRR}, abs/2010.11333.

\bibitem[{Wang et~al.(2021)Wang, Brovman, and
  Madhvanath}]{DBLP:journals/corr/abs-2102-06156}
Tian Wang, Yuri~M. Brovman, and Sriganesh Madhvanath. 2021.
\newblock \href {http://arxiv.org/abs/2102.06156} {Personalized embedding-based
  e-commerce recommendations at ebay}.
\newblock \emph{CoRR}, abs/2102.06156.

\bibitem[{Weischedel et~al.(2013)Weischedel, Palmer, Marcus, Hovy, Pradhan,
  Ramshaw, Xue, Taylor, Kaufman, Franchini et~al.}]{weischedel2013ontonotes}
Ralph Weischedel, Martha Palmer, Mitchell Marcus, Eduard Hovy, Sameer Pradhan,
  Lance Ramshaw, Nianwen Xue, Ann Taylor, Jeff Kaufman, Michelle Franchini,
  et~al. 2013.
\newblock Ontonotes release 5.0 ldc2013t19.
\newblock \emph{Linguistic Data Consortium, Philadelphia, PA}, 23.

\bibitem[{Yang and Katiyar(2020)}]{yang2020simple}
Yi~Yang and Arzoo Katiyar. 2020.
\newblock \href {http://arxiv.org/abs/2010.02405} {Simple and effective
  few-shot named entity recognition with structured nearest neighbor learning}.

\bibitem[{Yu et~al.(2018{\natexlab{a}})Yu, Guo, Yi, Chang, Potdar, Cheng,
  Tesauro, Wang, and Zhou}]{yu-etal-2018-diverse}
Mo~Yu, Xiaoxiao Guo, Jinfeng Yi, Shiyu Chang, Saloni Potdar, Yu~Cheng, Gerald
  Tesauro, Haoyu Wang, and Bowen Zhou. 2018{\natexlab{a}}.
\newblock \href {https://doi.org/10.18653/v1/N18-1109} {Diverse few-shot text
  classification with multiple metrics}.
\newblock In \emph{Proceedings of the 2018 Conference of the North {A}merican
  Chapter of the Association for Computational Linguistics: Human Language
  Technologies, Volume 1 (Long Papers)}, pages 1206--1215, New Orleans,
  Louisiana. Association for Computational Linguistics.

\bibitem[{Yu et~al.(2018{\natexlab{b}})Yu, Guo, Yi, Chang, Potdar, Cheng,
  Tesauro, Wang, and Zhou}]{DBLP:journals/corr/abs-1805-07513}
Mo~Yu, Xiaoxiao Guo, Jinfeng Yi, Shiyu Chang, Saloni Potdar, Yu~Cheng, Gerald
  Tesauro, Haoyu Wang, and Bowen Zhou. 2018{\natexlab{b}}.
\newblock \href {http://arxiv.org/abs/1805.07513} {Diverse few-shot text
  classification with multiple metrics}.
\newblock \emph{CoRR}, abs/1805.07513.

\bibitem[{Zhou et~al.(2018)Zhou, Khashabi, Tsai, and Roth}]{Zhou2018ZeroShotOE}
Ben Zhou, Daniel Khashabi, Chen-Tse Tsai, and Dan Roth. 2018.
\newblock Zero-shot open entity typing as type-compatible grounding.
\newblock In \emph{EMNLP}.

\end{thebibliography}
\bibliographystyle{acl_natbib}

\appendix

\section{Datasets Details}
\label{appendix:datasets}
\subsection{Statistics}
\label{appendix:datasets_stats}
Table \ref{tab:dataset_details} shows the statistics of original datasets we use in the main experiments.

\begin{table}[H]
\begin{center}
\begin{tabular}{lllc}
%\hline
\toprule
\bf Dataset & \bf Domain & \bf \# \bf Sent & \bf \# Labels \\
\midrule
Ontonotes & Mix & 76,714 & 18 \\
CoNLL'03 & News & 20,744 & 4 \\
WNUT'07 & Social & 5,690 & 6 \\
JNLPBA & Bio & 22,402 & 5 \\
NCBI-disease & Bio & 7,287 & 1 \\
I2B2'14 & Medical & 75,330 & 23 \\
\toprule
\end{tabular}
\end{center}
\caption{\label{tab:dataset_details} Original dataset statistics.}
\end{table}

\subsection{Label Names}
\label{appendix:datasets_label_names}

Table \ref{tab:dataset_labels} shows the original label names in each dataset and corresponding natural language forms we use in our experiments.
\begin{table}[H]
\small
\begin{center}
\begin{tabular}{l|c|l}
%\hline
\toprule
\bf Dataset & \makecell[c]{\bf Original \\ \bf Labels} & \makecell[c]{\bf Natural \\ \bf Language} \\
\midrule
\multirow{4}{*}{\bf CoNLL'03} & PER & person \\
& LOC & location \\
& ORG & organization \\
& MISC & miscellaneous \\
\hline
\multirow{18}{*}{\bf Ontonotes} & CARDINAL & cardinal \\
& DATE & date \\
& EVENT & event \\
& FAC & facility \\
& GPE & \makecell[l]{geographical social \\ political entity} \\
& LANGUAGE & language \\
& LAW & law \\
& LOC & location \\
& MONEY & money \\
& NORP & \makecell[l]{nationality religion \\ political} \\
& ORDINAL & ordinal \\
& ORG & organization \\
& PERCENT & percent \\
& PERSON & person \\
& PRODUCT & product \\
& QUANTITY & quantity \\
& TIME & time \\
& WORK\_OF\_ART & work of art \\
\hline
\multirow{6}{*}{\bf WNUT'17} & corporation & corporation \\
& creative-work & creative work \\
& group & group \\
& location & location \\
& person & person \\
& product & product \\
\hline
\multirow{4}{*}{\bf JNLPBA} & DNA & DNA \\
& RNA & RNA \\
& cell\_line & cell line \\
& cell\_type & cell type \\
& protein & protein \\
\hline
\makecell[l]{\bf NCBI- \\ \bf disease} & Disease & disease \\
\hline
\multirow{23}{*}{\bf I2B2'14} & AGE & age \\
& BIOID & biometric ID \\
& CITY & city \\
& COUNTRY & country \\
& DATE & date \\
& DEVICE & device \\
& DOCTOR & doctor \\
& EMAIL & email \\
& FAX & fax \\
& HEALTHPLAN & health plan number \\
& HOSPITAL & hospital \\
& IDNUM & ID number \\
& LOCATION\_OTHER & location \\
& MEDICALRECORD & medical record \\
& ORGANIZATION & organization \\
& PATIENT & patient \\
& PHONE & phone number \\
& PROFESSION & profession \\
& STATE & state \\
& STREET & street \\
& URL & url \\
& USERNAME & username \\
& ZIP & zip code \\
\bottomrule
\end{tabular}
\end{center}
\caption{\label{tab:dataset_labels} Original label names and their corresponding natural language formats.}
\end{table}

\section{Support Set Sampling Algorithm}
\label{apppendix:sampling_algorithm}
\begin{algorithm}
\small
\caption{Support set sampling}\label{alg:sampling}
\begin{algorithmic}[1]
\Require \# shot $K$, dataset $\mathcal{D}$, labels $\mathcal{L}_{\mathcal{D}}$
\State Initialize support set $\mathcal{S}$=\{\}, Count\textsubscript{$\ell_{i}$}=0\ ($\forall \ell_{i} \in \mathcal{L_D}$)
\For{$\ell$ in $\mathcal{L_D}$}
    \While{Count\textsubscript{$\ell$} < $K$}
        \State Randomly pick $(\boldsymbol{t}, \boldsymbol{y})$ from $\mathcal{D} \setminus \mathcal{S}$ that $\boldsymbol{y}$ include $\ell$
        \State $\mathcal{S} \gets \mathcal{S} \cup (\boldsymbol{t}, \boldsymbol{y}) $
        \State Update all Count\textsubscript{$\ell_{i}$}\ ($\forall \ell_{i} \in \mathcal{L_D}$)
    \EndWhile
\EndFor
\For{$(\boldsymbol{t}, \boldsymbol{y})$ in $\mathcal{S}$}
    \State $\mathcal{S} = \mathcal{S} \setminus (\boldsymbol{t}, \boldsymbol{y})$
    \State Update all Count\textsubscript{$\ell_{i}$}\ ($\forall \ell_{i} \in \mathcal{L_D}$)
    \If{Any Count\textsubscript{$\ell_{i}$} < $K$}
        \State $\mathcal{S} = \mathcal{S} \cup (\boldsymbol{t}, \boldsymbol{y})$
        \State Update all Count\textsubscript{$\ell_{i}$}\ ($\forall \ell_{i} \in \mathcal{L_D}$)
    \EndIf
\EndFor
\end{algorithmic}
\end{algorithm}

\section{Hardware for Experiments}
\label{appendix:hardware}
We provide details about hardware we use to produce numbers for each baseline models. We run experiments for Struct NN shot model on NVIDIA V100 GPU with 32GB of memory, while for all other models (including baselines and our models) we use NVIDIA V100 GPU with 16GB of memory. 

\section{Visualization of Results}
\label{appendix:visualization}
\begin{figure*}
    \centering \includegraphics[scale=0.25]{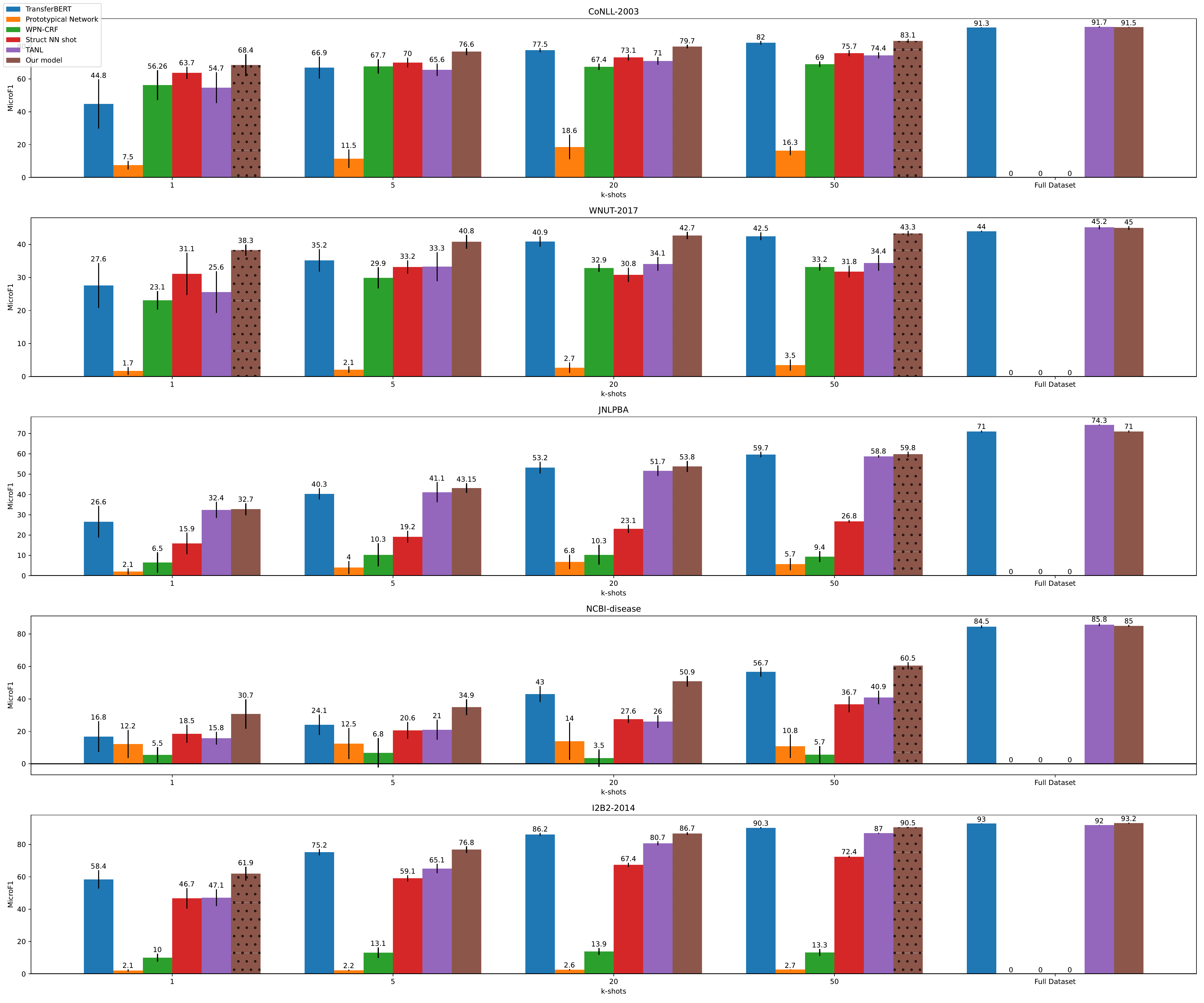}
    \captionof{figure}{Visualization of the results in Table \ref{tab:results}. Results on test set of all datasets. All numbers indicate micro F1 scores except noted otherwise. Results for low resource settings are average of 10 runs with different support set sampling. Results for high resource setting are average of 5 runs with different random seeds. For some baselines we cannot run the released implementation from originally papers due to GPU out of memory and they are marked as 0.}
    \label{fig:main_results_bar_chart}
\end{figure*}

We visualize the results in Table \ref{tab:results} with bar chart, as shown in Figure \ref{fig:main_results_bar_chart}.

\section{Contextualized Label Representations}
\label{appendix:contextual_label_rep}

In this experiment, we compute contextualized label representations by randomly selecting a sentence from the support set that contains an entity of the type, and replace that entity with the label name in the sentence. We encode this sentence with the label encoder and compute the average pooling as the label representation. The label names used are in their natural language form with BIO schemes per \ref{model:architecture}. We depict this process in Figure \ref{fig:lsvscls}. At inference time, to avoid biasing toward any particular sentence, we randomly choose 10 sentences from the support set for each label and average their representations as the final label representations.\footnote{When there are less than 10 sentences for a given label in the support set, we use all the available sentences. Sentences are selected once then fixed. We also experimented by randomly choosing one fixed sentence for both training and inference from the support set, but preliminary results show it is worse than our current method.}

\begin{center}
  \includegraphics[scale=0.7]{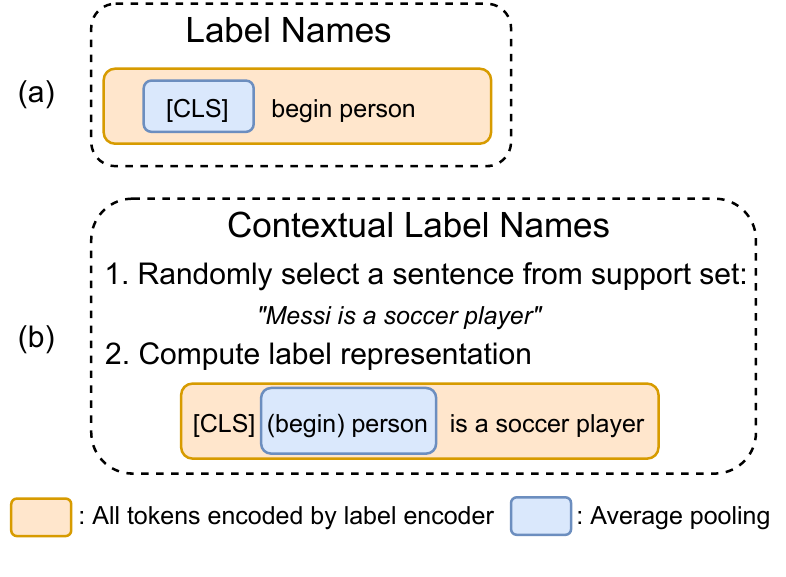}
  \captionof{figure}{Differences between contextualized label representations and label representations in isolation.}
  \label{fig:lsvscls}
\end{center}

We perform experiments on FEW-NERD dataset \cite{DBLP:journals/corr/abs-2105-07464}.\footnote{As in the other experiments, we pre-finetune all models on Ontonotes then continue finetuning on target datasets.} This dataset consists of 8 coarse-grained and 66 fine-grained entity types in hierarchy. The fine-grained entity types under the same coarse-grained type are semantically close.

 Results are shown in Table \ref{tab:contexual_results} and Appendix \ref{appendix:contextual_label_rep}. In the following, we show 1-shot results under ``Person'' coarse-grained type for FEW-NERD dataset.\footnote{Fine-grained entity types under ``Person'' are: Actor, Artist/author, Athlete, Director, Politician, Scholar and Soldier.}  By using contextual label names, we observe a decrease in model performance by 3.5 F1 points on FEW-NERD, compared to when only label names are used. This suggests that the trained label encoder is capable of capturing critical semantics with only label names, even without contexts to help distinguish semantically close labels. % However, the model with contextual label names

\begin{table}[!ht]
\small
\begin{center}
\begin{tabular}{lccccc}
%\hline
\toprule
\multirow{2}{*}{\bf Datasets} & \multicolumn{2}{c}{\bf Model} \\
\cline{2-3}
 & \bf \makecell[c]{Ours} & \bf Ours + context  \\
\toprule
CoNLL'03 & 69.0$\pm$6.9 & 70.8$\pm$4.1 \\
%\midrule
WNUT17 & 48.2$\pm$1.7 & 51.8$\pm$1.8 \\
%\midrule
JNLPBA & 31.5$\pm$2.9 & 30.1$\pm$3.2  \\
%\midrule
FEW-NERD-Person & 32.5$\pm$8.1 & 29.0$\pm$7.1 \\
\toprule
\end{tabular}
\end{center}
\caption{\label{tab:contexual_results} 1-shot micro F1 on development set across various datasets and models. Ours: Our model with label names. Ours+context: Our model with contextual label names. Numbers are averaged across 10 different random samplings.}
\end{table}

\subsection{Additional Experiment 1}

We present additional experiments on contextual label representations. We will first introduce more details on the FEW-NERD dataset, then describe methods we explore to contextualize labels, finally we will show experiment results.
% To evaluate our model on fine-grained entity types, we construct a dataset FEW-NERD-person from FEW-NERD dataset. FEW-NERD-person has sentences that contains fine-grained person subtypes - actor, author, athlete, director, politician, scholar and soldier. 
% The dataset contains 8,357 sentences and 7 labels.
To validate whether contextual label representation can improve model performance in scenarios where labels are semantically close, we perform experiments on one additional dataset: FEW-NERD \cite{DBLP:journals/corr/abs-2105-07464}. FEW-NERD is a human annotated NER dataset that consists of 188,238 sentences. It has a hierarchy of 8 coarse-grained and 66 fine-grained entity types. The fine-grained entity types under each coarse-grained type are usually semantically close. All sentences are sourced from Wikipedia. We use train/dev/test split from the original dataset distribution.

We select ``Person'' and ``Art'' coarse-grained entity types for the experiments, because we think fine-grained entity types under them have closest semantic similarities. Specifically, we take one coarse-grained entity type at a time, and remove all entity annotations that do not belong to it, on train, dev and test split. After removal, comparing with the original dataset, the resulting dataset has much more sentences with no annotation than sentences that have at least one annotations. To mitigate this entity distribution shifting, we randomly remove sentences that do not contain any annotations, such that the resulting dataset has the same percentage of sentences with annotations as the original dataset. We perform this process on ``Person'' and ``Art'' types and result in two datasets called ``FEW-NERD-Person'' and ``FEW-NERD-Art''. The statistics for these two datasets are shown in Table \ref{appendix:few_nerd_stats}. The original entity types and their corresponding natural language format are shown in Table \ref{appendix:few_nerd_labels}

\begin{table*}[t]
% \small
\begin{center}
\begin{tabularx}{0.75\textwidth}{lcccccc}
%\hline
\toprule
\multirow{2}{*}{\bf Dataset} & \multirow{2}{*}{\bf \# Labels} & \multicolumn{4}{c}{\bf Support Set Shot} & \multirow{2}{*}{\bf Dev}\\
\cline{3-6}
 & & \bf 1 & \bf 5 & \bf 20 & \bf 50 & \\
\toprule
FEW-NERD-Person & 7 & 19.0 & 66.7 & 212.7 & 508.9 & 4437.0 \\
FEW-NERD-Art & 5 & 41.5 & 123.5 & 412.2 & 2569.0 & 1364.0 \\
\toprule
\end{tabularx}
\end{center}
\caption{\label{appendix:few_nerd_stats} Number of sentences in support set and dev set for FEW-NERD-Person and FEW-NERD-Art datasets. Numbers are averaged across 10 different random samplings.}
\end{table*}

\begin{table}[H]
\small
\begin{center}
\begin{tabular}{l|c|l}
%\hline
\toprule
\bf Dataset & \makecell[c]{\bf Original \\ \bf Labels} & \makecell[c]{\bf Natural \\ \bf Language} \\
\midrule
\multirow{7}{*}{\makecell[c]{\bf FEW-NERD-\\ \bf Person}} & person-actor & actor \\
& person-artist/author & artist author \\
& person-athlete & athlete \\
& person-director & director \\
& person-politician & politician \\
& person-scholar & scholar \\
& person-soldier & soldier \\
\hline
\multirow{5}{*}{\makecell[c]{\bf FEW-NERD-\\ \bf Art}} & art-broadcastprogram & \makecell[l]{broadcast-\\ program} \\
& art-film & film \\
& art-music & music \\
& art-painting & painting \\
& art-writtenart & written art \\
\bottomrule
\end{tabular}
\end{center}
\caption{\label{appendix:few_nerd_labels} Original label names and their corresponding natural language formats for FEW-NERD-Person and FEW-NERD-Art datasets.}
\end{table}

\subsection{Additional Experiment 2}

In this experiment, we replace the entity in the selected sentence with different texts rather than label names.

We experiment with various schemes for the new span and use the following terminology to describe them. \textit{TOKEN} refers to the original token that is replaced. \textit{LABEL} refers to the label name that the token is annotated with. \textit{BIO-TAG} refers to the natural BIO tag that the token is annotated with. For the example illustrated in Figure \ref{fig:lsvscls}, \textit{TOKEN} corresponds to "Messi", \textit{LABEL} corresponds to "person", \textit{BIO-TAG} corresponds to "begin". We hypothesize that the \textit{TOKEN} gives natural context to the labels since it is unmodified sentence, \textit{LABEL} captures the semantic information in label names and \textit{BIO-TAG} helps differentiate the B and I chunks for the label. In addition, we experiment to replace the entity with "[MASK]" token to make the label reprensetation close to BERT pretraining inputs. The various schemes are illustrated with example in Figure \ref{fig:cls_variations}.

\begin{center}
  \includegraphics[scale=0.8]{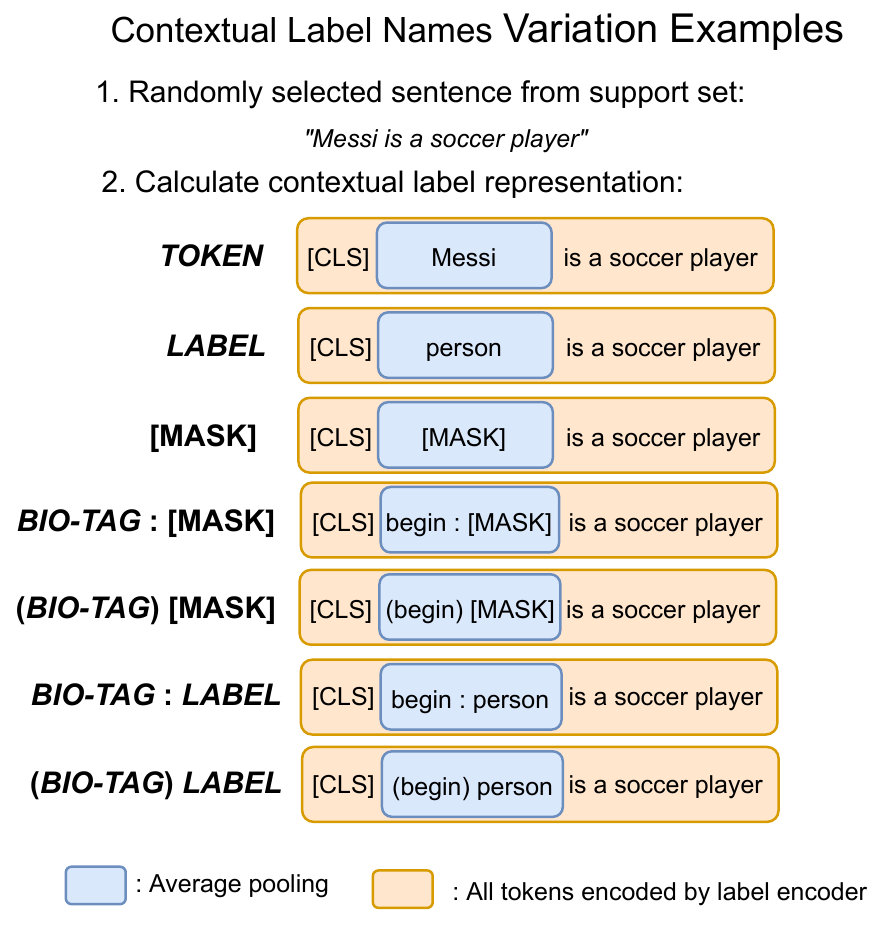}
  \captionof{figure}{Example for contextual label representation. }
  \label{fig:cls_variations}
\end{center}

\subsection{Results}
The results from various schemes of the new span is compared with TransferBERT and our model which encodes label names only. This is summarized in Table \ref{tab:contextual_results}. 

\textit{TOKEN} scheme is the simplest way to get a contextualized representation of a label where we pool the representations of all the tokens annotated with the label.  Although performance of this scheme is better than TransferBERT, comparing with other schemes, we see that this model performs poorly. Here no new information is added to the model and the text that the label encoder and document encoder encodes is similar. In order to provide our model prior knowledge about the label name from BERT encoder, we use \textit{LABEL} scheme. We see that this scheme performs better than \textit{TOKEN} across datasets suggesting that the prior knowledge about label semantics helps to improve performance. 

One limitation with \textit{LABEL} scheme is that the replaced token is same for both B and I chunks in BIO scheme. For example, to get contextualized representation for B-PER in the document  "Lionel Messi is a soccer player", the document will be transformed to "person person is a soccer player", where B and I chunks are confused. "\textit{BIO-TAG} : \textit{LABEL}" scheme addresses this by prefixing the natural language BIO chunk name to the label name. We see improvements in performance compared with \textit{LABEL} scheme.

% A downside to prefixing the natural language BIO chunk name is that this sometimes breaks the structure of the sentence, for example,  "begin person is a soccer player". We hypothesize that this is very different from what BERT has been pre-trained on and impacts the performance. To validate this, we run experiments on "(\textit{BIO-TAG}) \textit{LABEL}" scheme, which resembles natural language better. We observe that this scheme performs the best. 

When we incorporate the ``[MASK]'' token  from BERT pretraining, we find that this does not perform as well as other schemes that contains label names. This further prove that semantics in label names is critical.
% We also analyze another line of idea where we try to incorporate the information learnt by BERT's [MASK] token during its pre-training in the representation of the label. The schemes "[MASK]", "\textit{BIO-TAG} : [MASK]" and "(\textit{BIO-TAG}) [MASK]" replaces the annotation with new spans containing [MASK] token. As we see from Table \ref{tab:contextual_results} this does not perform as well as the schemes that use the label name itself. This suggests that the label semantics encoded by the label name helps in model performance.

\begin{table*}[t]
\centering
\small
\setlength{\tabcolsep}{5pt}
\vspace{-.3cm}
%       & {\bf 1 Shot} & {\bf 5 Shot} & {{\bf 20 Shot}} & {{\bf 50 Shot}} \\
%\toprule

\begin{tabularx}{\textwidth}{p{\sidedescwidth} p{\modelcolwidth} *{5}{C}}
\topleftdesc{CoNLL03}{12} 
& {\bf 1 Shot} & {\bf 5 Shot} & {{\bf 20 Shot}} & {{\bf 50 Shot}} \\
\toprule
& TransferBERT & 47.6 $\pm$ 15.5  & 69.9 $\pm$ 6.0  & 80.1 $\pm$ 1.7  & 85.1 $\pm$ 1.1 \\
& Ours, label name only & 69.0 $\pm$ 6.9  & \bf 78.6 $\pm$ 1.8  & \bf 82.1 $\pm$ 1.5  & \bf 85.9 $\pm$ 1.2 \\
\cmidrule(lr){2-7}
& \textit{TOKEN} & 60.1 $\pm$ 16.8  & 75.0 $\pm$ 4.2  & 80.0 $\pm$ 1.8  & 84.3 $\pm$ 1.1 \\
& \textit{LABEL} & 61.4 $\pm$ 12.7  & 74.2 $\pm$ 2.9  & 80.4 $\pm$ 1.9  & 84.6 $\pm$ 1.2 \\
& [MASK] & 61.2 $\pm$ 6.1  & 72.9 $\pm$ 5.8  & 81.5 $\pm$ 2.2  & 85.3 $\pm$ 0.9 \\
& \textit{BIO-TAG} : [MASK] & 60.8 $\pm$ 15.4  & 74.5 $\pm$ 5.6  & 81.3 $\pm$ 1.5  & 85.2 $\pm$ 0.8 \\
& (\textit{BIO-TAG}) [MASK] & 66.8 $\pm$ 6.7  & 74.6 $\pm$ 7.0  & 81.6 $\pm$ 1.8  & 85.3 $\pm$ 1.0 \\
& \textit{BIO-TAG} : \textit{LABEL} & 69.2 $\pm$ 6.4  & 76.1 $\pm$ 2.1  & 80.8 $\pm$ 1.9  & 84.9 $\pm$ 1.1 \\
& (\textit{BIO-TAG}) \textit{LABEL} & \bf 70.8 $\pm$ 4.2  & 76.5 $\pm$ 1.6  & 81.2 $\pm$ 2.0  & 84.7 $\pm$ 1.1 \\
\end{tabularx}

\begin{tabularx}{\textwidth}{p{\sidedescwidth} p{\modelcolwidth} *{5}{C}}
\topleftdesc{WNUT17}{12} \\
\toprule
& TransferBERT & 35.6 $\pm$ 11.2  & 44.7 $\pm$ 5.6  & 50.3 $\pm$ 1.7  & 51.7 $\pm$ 1.9 \\
& Ours, label name only & 48.3 $\pm$ 1.7  & 51.2 $\pm$ 1.4  & 53.2 $\pm$ 1.1  & 54.1 $\pm$ 1.3 \\
\cmidrule(lr){2-7}
& \textit{TOKEN} & 42.8 $\pm$ 12.3  & 49.9 $\pm$ 1.9  & 53.1 $\pm$ 1.8  & 53.9 $\pm$ 1.8 \\
& \textit{LABEL} & 48.9 $\pm$ 3.0  & 51.4 $\pm$ 2.1  & 53.0 $\pm$ 1.6  & 53.9 $\pm$ 1.5 \\
& [MASK] & 45.0 $\pm$ 3.5  & 47.1 $\pm$ 2.2  & 50.2 $\pm$ 2.3  & 51.9 $\pm$ 1.6 \\
& \textit{BIO-TAG} : [MASK] & 46.8 $\pm$ 2.8  & 49.6 $\pm$ 1.7  & 51.3 $\pm$ 2.8  & 52.7 $\pm$ 1.0 \\
& (\textit{BIO-TAG}) [MASK] & 45.6 $\pm$ 4.8  & 48.5 $\pm$ 2.6  & 51.2 $\pm$ 2.7  & 52.6 $\pm$ 1.7 \\
& \textit{BIO-TAG} : \textit{LABEL} & 51.2 $\pm$ 2.2  & 52.6 $\pm$ 1.8  & 53.6 $\pm$ 1.4  & \bf 54.8 $\pm$ 0.6 \\
& (\textit{BIO-TAG}) \textit{LABEL} & \bf 51.9 $\pm$ 1.8  & \bf 52.3 $\pm$ 1.2  & \bf 53.7 $\pm$ 1.5  & 54.0 $\pm$ 1.3 \\
\end{tabularx}

\begin{tabularx}{\textwidth}{p{\sidedescwidth} p{\modelcolwidth} *{5}{C}}
\topleftdesc{NCBI-diseas}{12} \\
\toprule
& TransferBERT & 15.1 $\pm$ 9.4  & 19.5 $\pm$ 6.0  & 37.0 $\pm$ 4.1  & 51.2 $\pm$ 4.1 \\
& Ours, label name only & \bf 31.4 $\pm$ 9.2  & \bf 30.2 $\pm$ 4.3  & \bf 45.8 $\pm$ 3.4  & \bf 57.3 $\pm$ 2.6 \\
\cmidrule(lr){2-7}
& \textit{TOKEN} & 18.7 $\pm$ 10.3  & 22.5 $\pm$ 6.4  & 40.9 $\pm$ 5.6  & 53.8 $\pm$ 4.1 \\
& \textit{LABEL} & 26.9 $\pm$ 8.3  & 28.7 $\pm$ 4.2  & 40.2 $\pm$ 3.7  & 52.3 $\pm$ 2.9 \\
& [MASK] & 18.1 $\pm$ 9.6  & 22.2 $\pm$ 4.0  & 38.2 $\pm$ 5.3  & 53.0 $\pm$ 4.0 \\
& \textit{BIO-TAG} : [MASK] & 17.7 $\pm$ 10.0  & 22.3 $\pm$ 4.2  & 40.0 $\pm$ 4.5  & 52.1 $\pm$ 3.7 \\
& (\textit{BIO-TAG}) [MASK] & 17.5 $\pm$ 11.5  & 23.6 $\pm$ 4.1  & 38.8 $\pm$ 4.7  & 51.9 $\pm$ 4.0 \\
& \textit{BIO-TAG} : \textit{LABEL} & 26.8 $\pm$ 7.4  & 26.2 $\pm$ 3.8  & 42.0 $\pm$ 4.1  & 54.4 $\pm$ 3.4 \\
& (\textit{BIO-TAG}) \textit{LABEL} & 26.8 $\pm$ 9.2  & 26.7 $\pm$ 3.3  & 43.9 $\pm$ 3.8  & 54.6 $\pm$ 3.3 \\
\end{tabularx}

\begin{tabularx}{\textwidth}{p{\sidedescwidth} p{\modelcolwidth} *{5}{C}}
\topleftdesc{JNLPBA}{12} \\
\toprule
& TransferBERT & 26.3 $\pm$ 8.0  & 41.8 $\pm$ 3.0  & \bf 55.9 $\pm$ 3.5  & \bf 64.3 $\pm$ 1.3 \\
& Ours, label name only & \bf 31.5 $\pm$ 3.0 & \bf 43.3 $\pm$ 2.8  & 55.8 $\pm$ 3.4  & 63.6 $\pm$ 1.0 \\
\cmidrule(lr){2-7}
& \textit{TOKEN} & 29.0 $\pm$ 6.5  & 43.2 $\pm$ 2.4  & \bf 55.9 $\pm$ 3.6  & 63.8 $\pm$ 1.2 \\
& \textit{LABEL} & 28.4 $\pm$ 4.3  & 40.8 $\pm$ 2.5  & 54.3 $\pm$ 3.4  & 62.5 $\pm$ 1.3 \\
& [MASK] & 25.4 $\pm$ 6.5  & 36.5 $\pm$ 2.2  & 51.0 $\pm$ 3.7  & 60.2 $\pm$ 1.5 \\
& \textit{BIO-TAG} : [MASK] & 24.9 $\pm$ 5.1  & 36.0 $\pm$ 2.5  & 50.5 $\pm$ 4.2  & 60.5 $\pm$ 1.7 \\
& (\textit{BIO-TAG}) [MASK] & 24.8 $\pm$ 6.5  & 37.1 $\pm$ 2.9  & 50.4 $\pm$ 4.1  & 60.3 $\pm$ 1.7 \\
& \textit{BIO-TAG} : \textit{LABEL} & 30.4 $\pm$ 4.6  & 41.9 $\pm$ 2.5  & 55.5 $\pm$ 3.3  & 62.9 $\pm$ 1.1 \\
& (\textit{BIO-TAG}) \textit{LABEL} & 30.1 $\pm$ 3.2  & 41.4 $\pm$ 2.2  & 55.1 $\pm$ 3.2  & 62.8 $\pm$ 1.5 \\
\end{tabularx}

\begin{tabularx}{\textwidth}{p{\sidedescwidth} p{\modelcolwidth} *{5}{C}}
\topleftdesc{FN-Person}{7} \\
\toprule
& TransferBERT & 13.2 $\pm$ 5.0  & 24.0 $\pm$ 7.4  & 48.7 $\pm$ 3.4  & 66.9 $\pm$ 3.0 \\
& Ours, label name only & \bf 32.5 $\pm$ 8.1  & \bf 51.0 $\pm$ 7.0  & \bf 66.2 $\pm$ 2.0  & \bf 72.0 $\pm$ 0.7 \\
\cmidrule(lr){2-7}
& (\textit{BIO-TAG}) \textit{LABEL} & 29.0 $\pm$ 7.2  & 50.6 $\pm$ 6.3  & \bf 66.2 $\pm$ 2.0  & 71.2 $\pm$ 0.9 \\
\end{tabularx}

\begin{tabularx}{\textwidth}{p{\sidedescwidth} p{\modelcolwidth} *{5}{C}}
\topleftdesc{FN-Art}{6} \\
\toprule
& TransferBERT & 19.4 $\pm$ 10.9  & 43.1 $\pm$ 9.8  & 69.5 $\pm$ 1.7  & 98.9 $\pm$ 0.3 \\
& Ours, label name only & \bf 44.5 $\pm$ 8.8  & \bf 56.3 $\pm$ 4.6  & \bf 70.5 $\pm$ 1.8  & \bf 99.1 $\pm$ 0.1 \\
\cmidrule(lr){2-7}
& (\textit{BIO-TAG}) \textit{LABEL} & 41.3 $\pm$ 10.8  & 56.0 $\pm$ 3.8  & 69.4 $\pm$ 2.0  & 98.9 $\pm$ 0.2 \\
\toprule
\end{tabularx}

\caption{Results on development set across all datasets. FN-Person = FEW-NERD-Person. FN-Art = FEW-NERD-Art. All numbers indicate micro F1 scores and are average of 10 runs with different support set sampling.}
\label{tab:contextual_results}

\end{table*}
% srikad replace

\end{document}